\documentclass{article}
\usepackage{preprint,times}
\usepackage[utf8]{inputenc}
\usepackage[T1]{fontenc}
\usepackage{amsmath,amssymb,booktabs,graphicx,multirow}
\usepackage{natbib}
\usepackage{xcolor}
\usepackage{hyperref}
\hypersetup{pdftitle={Computation, Not Memory: Chain-of-Thought and State Tracking in LLM Agents},pdfauthor={Pedram Mouseli, Massieh Moayedi},colorlinks=true,linkcolor=[rgb]{0.1,0.2,0.5},citecolor=[rgb]{0.1,0.2,0.5},urlcolor=[rgb]{0.1,0.2,0.5}}
\usepackage{microtype}
\usepackage{placeins}
\newcommand{\cmd}[1]{\texttt{#1}}
\newcommand{\cond}[1]{\textsf{#1}}
\graphicspath{{figures/}}
\title{Computation, Not Memory: Chain-of-Thought\\and State Tracking in LLM Agents}
\author{Pedram Mouseli\thanks{Corresponding author: \texttt{pedram.mouseli@utoronto.ca}} \quad Massieh Moayedi\\
University of Toronto}
\date{}

\begin{document}
\maketitle

\lhead{Preprint}

\begin{abstract}
Language-model agents often reason before they act, and the value of monitoring that reasoning depends on how it relates to their actions. We ask how an agent's actions depend on its reasoning when the correct action depends on a changing state. In a small file-system task, background events announce which file's lock has been toggled but not its new value. The agent therefore has to reconstruct the current locks from its history before it can move files safely. With Qwen3-8B and Qwen3.5-9B, we separate two roles of reasoning: generating it before each action, and retaining earlier reasoning in the prompt. Most of the reduction in lock-violating commands comes from reasoning written at the moment of action. Retaining earlier reasoning adds little when the agent also reasons at each step, but it helps an agent that does not reason, and it helps in longer episodes. Retained notes have a limit: when a later event makes a note wrong, an agent without current reasoning responds little, while current reasoning makes the action follow the event. The benefit depends on what the reasoning says. Filler text, reasoning written for another history and an accurate table of the state do not reproduce it. When we give the agent reasoning written for a history that differs in one lock value, the agent acts on that reasoning even against its own history. For Qwen3-8B, this holds also when that value has since been toggled. Without reasoning, state readouts are less reliable after toggles, and Qwen3-8B's answers about the state favor the outdated value. The state answer and the action also depend on different parts of the prompt. In this setting, the benefit of reasoning for rule-following comes mainly from generating it at each step, where it makes the action respond to later events. The state that can be read out without reasoning is a limited guide to how the agent acts.
\end{abstract}

\section{Introduction}
 \enlargethispage{2\baselineskip}

Language models are increasingly deployed as agents that reason before they act \citep{yao2023react}, and their earlier interaction usually remains in the prompt. Their written reasoning is also a potential source of information about their decisions, which motivates proposals to monitor agents through their chains of thought \citep{baker2025monitoring,korbak2025monitorability}. How useful the reasoning is for this purpose depends on the role it plays in the agent's decisions. In theory, writing intermediate tokens extends the serial computation that a transformer can carry out before it answers \citep{merrill2024expressive,li2024serial,brown-cohen2026opaque}. In practice, the role of reasoning in a given agent and task is an empirical question. A trace may carry out computation that the next action needs, leave notes that later decisions can use, or describe a decision without causing it. Telling these roles apart requires interventions on when reasoning is generated, on what is retained, and on what the trace says.

We study these roles in a task that requires keeping track of a changing state. We built a small file-system environment in which files are locked or unlocked. Background events toggle the locks and announce which file changed, but not its new value. Safe moves depend on the current lock state, which is determined by the history and must be reconstructed from it. The full history of observations and executed commands stays in the prompt in every condition, so the evidence needed to reconstruct each lock state is always available. An agent can still fail to use this evidence. A state change also adds a requirement: an earlier conclusion must be revised when a later event changes the fact it was based on.

Reasoning can help such an agent in two ways. \emph{Generating} reasoning before an action gives the model extra computation at the moment of decision. \emph{Retaining} earlier reasoning in the prompt, which we call \emph{persistence}, gives it notes that it could consult instead of working out the state again \citep{lanchantin2023selfnotes}. The chat templates of Qwen3 and Qwen3.5 separate the two by default: the agent reasons at every turn, but earlier reasoning is removed from the prompt \citep{qwen3modelcard,qwen35modelcard}.

We ask four questions, each building on the previous one.
\begin{enumerate}
\item \textbf{Does the benefit come from generating reasoning or from retaining it?} Generating reasoning reduces lock-violating commands by a large margin in both models. Retaining earlier reasoning adds little when the agent also reasons at each step of an eight-turn episode. It does help an agent that does not reason, and it helps in sixteen-turn episodes (Section~\ref{sec:genpersist}).
\item \textbf{Why does retained reasoning not replace current reasoning?} A retained note records the state at the time it was written. When a later toggle makes a correct note wrong, an agent without current reasoning responds little, while current reasoning makes the action follow the event. Labelling the order of events helps Qwen3-8B further when it reasons (Section~\ref{sec:fresh}).
\item \textbf{Does the action use what the current reasoning says?} Neither filler of the same length nor an accurate state table reproduces the benefit of the model's own reasoning. When we swap reasoning between histories, the action follows the reasoning rather than the history, for Qwen3-8B also after a toggle (Sections~\ref{sec:fixed} and~\ref{sec:cf}).
\item \textbf{Without reasoning, does a readout of the state explain the action?} State readouts become less reliable after toggles, and Qwen3-8B's answers about the state favor the outdated value, much as a retained note does. An accurate state does not ensure a safe direct action. The state answer and the action also depend on different positions (Section~\ref{sec:causal}).
\end{enumerate}
Together, the answers form one picture. Retained notes go out of date when the state changes, and readouts of the state become less reliable. Reasoning at the current decision makes the action respond to later events, and the action follows what this reasoning says. This is the sense of our title: when the agent reasons at each step, the benefit comes from computation at the current decision rather than from reasoning kept as memory.

\section{Related work}
\label{sec:related}

\paragraph{Reasoning as computation and as memory.} Chain-of-thought prompting \citep{wei2022chain,yao2023react} improves multi-step problem solving, and intermediate tokens increase what a transformer can compute \citep{merrill2024expressive,li2024serial}. Filler or pause tokens can sometimes replace the content of the reasoning \citep{pfau2024dot,goyal2024pause}. Other work uses generated text as memory: Self-Notes interleave notes with the input \citep{lanchantin2023selfnotes}, and agent systems keep reflections or external memories across steps \citep{shinn2023reflexion,park2023generative,packer2023memgpt}. We measure the two roles separately while the full event history remains available, and we test what happens to retained notes when the state changes after they were written.

\paragraph{Faithfulness and monitoring.} Reasoning traces do not always reflect the computation behind an answer \citep{turpin2023unfaithful,chen2025reasoning}, and answers can sometimes be decoded before the reasoning begins \citep{cox2026posthoc}. The value of monitoring reasoning may therefore depend on how much computation must pass through it \citep{korbak2025monitorability,brown-cohen2026opaque}. Truncating the reasoning, inserting mistakes or replacing it with filler tests whether an answer depends on it \citep{lanham2023faithfulness}, and resampling measures the influence of single steps \citep{bogdan2025thoughtanchors,macar2025thoughtbranches}. Our controls, swaps and edits apply this kind of test to an agent whose state and actions can be checked against the environment.

\paragraph{State in activations.} Board states can be decoded, sometimes linearly, from models trained on game moves \citep{li2023emergent,nanda2023emergent}. Language models also track entity states that change through sequences of operations \citep{li2021implicit,kim2023entity,li2025track}, bind attributes to entities \citep{feng2024binding}, and implement entity tracking with identifiable circuits \citep{prakash2024finetuning}. High probe accuracy does not show that a model uses the decoded information \citep{hewitt2019control,belinkov2022probing,elazar2021amnesic}. Activation patching tests which activations affect an output \citep{meng2022locating,zhang2024patching,heimersheim2024patching,geva2023dissecting}, and difference-of-means directions tend to have larger causal effects than logistic-regression directions \citep{marks2024geometry}. Steering and subspace patching require care in interpretation \citep{turner2023steering,tan2024steering,makelov2024subspace}.

\section{Experimental setup}
\label{sec:setup}

\begin{figure}[t]
\centering
\includegraphics[width=0.88\linewidth]{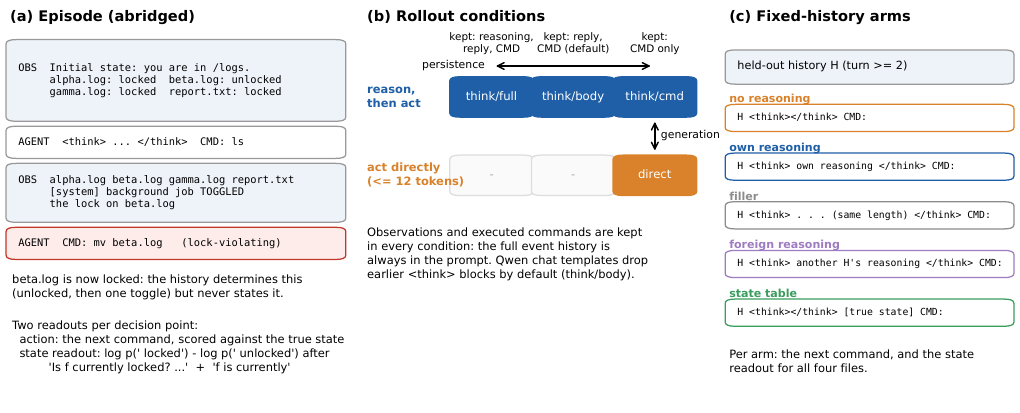}
\caption{\textbf{Design} (schematic). (a)~An abridged episode; toggles are announced without their result. (b)~Rollout conditions: whether the agent reasons (rows) and which part of each earlier reply is retained (columns). (c)~Fixed-history arms: one saved prompt completed with different thought blocks.}
\label{fig:design}
\end{figure}

\paragraph{Environment and task.} The file system has three log files (\cmd{alpha.log}, \cmd{beta.log}, \cmd{gamma.log}) and a goal file (\cmd{report.txt}). Each file is either in \cmd{/logs} or in \cmd{/archive}, and each is either locked or unlocked. At each turn the agent issues one command (\cmd{ls}, \cmd{cd}, \cmd{stat}, \cmd{lock}, \cmd{unlock}, \cmd{mv}, \cmd{rm} or \cmd{done}). The task is to archive every unlocked log file, never to touch a locked file, and to leave \cmd{report.txt} locked at the end. The first observation lists the lock value of every file. After the agent's second, fourth and sixth commands, a background job toggles the lock of a random file and announces that it did so, but not the new value (Figure~\ref{fig:design}a). The \cmd{lock} and \cmd{unlock} commands reply only ``ok''. As a result, the current lock state is always determined by the visible history, and a symbolic replay of the history recovers it exactly. The agent, however, has to combine events across turns to know the current state. An episode has at most eight turns, and the agent can end it with \cmd{done} from the seventh turn on (Appendix~\ref{app:env}).

\paragraph{Models and formats.} We use two open-weight models of similar size from different generations: Qwen3-8B, a transformer with standard attention, and Qwen3.5-9B, which combines linear and full attention \citep{yang2025qwen3,qwen35}. Each model can either reason in a thought block or answer directly, within the same checkpoint, so we can switch reasoning on and off without changing the weights. We run both models in their chat format. Appendix~\ref{app:env} also reports Qwen3-8B-Base with a plain-text prompt.

\paragraph{Generation and persistence.} In the \emph{reasoning} mode, the model writes a thought block and then a reply that ends with a command. In the \emph{direct} mode, reasoning is disabled, the prompt ends with \cmd{CMD:}, and the model writes only the command. The persistence setting decides which part of each earlier reply is retained in later prompts. With \cond{full}, the reasoning and the reply are retained. With \cond{body}, the default of the chat templates, only the reply is retained; we call the retained reply text the agent's \emph{notes}. With \cond{cmd}, only the executed command is retained. Observations and executed commands are retained in every condition, so \cond{cmd} removes earlier reasoning but not the event history. We name conditions by mode and persistence, for example \cond{think/cmd} or \cond{direct/cmd}, and use the same names when we vary the retained text of saved histories. Let $V(c)$ denote the violation rate of condition $c$ (defined below). We measure the effect of generating reasoning and the effect of retaining it by
\begin{equation}
\Delta_{\text{gen}} = V(\cond{direct/cmd}) - V(\cond{think/cmd}), \qquad
\Delta_{\text{ret}} = V(\cond{think/cmd}) - V(\cond{think/full}).
\label{eq:contrasts}
\end{equation}
Neither condition in $\Delta_{\text{gen}}$ retains earlier reasoning, and the two conditions in $\Delta_{\text{ret}}$ both generate it. $\Delta_{\text{gen}}$ compares two interfaces, with and without a thought block, and does not hold the amount of computation fixed. We run 500 episodes per condition for Qwen3-8B and 300 for the other models, and sample at temperature 0.7. Episode $i$ starts from the same initial state in every condition. Each reasoning reply has a token budget (4{,}000 tokens for all Qwen3 chat runs; Appendix~\ref{app:repro}). If a reply reaches the budget before the thought block is closed, we close the block and then elicit the command. This \emph{forced closure} occurs on 2.0--6.9\% of reasoning turns in the main runs, and excluding these turns leaves the effects unchanged (Appendix~\ref{app:behaviour}).

\paragraph{Saved histories.} Rollouts compare agents that end up in different states. We therefore also complete saved prompts under controlled changes, using greedy decoding, which always selects the most probable next token. In the \emph{fixed-history} experiment, we take 400 prompts from held-out \cond{think/body} episodes (third turn or later) and complete each with five versions of the thought block, which we call arms (Figure~\ref{fig:design}c). The arms are an empty block (\emph{no reasoning}), the model's own reasoning, and filler with as many dots as the own reasoning has tokens. Two further arms use the reasoning the model wrote for a prompt from another episode (\emph{other-history reasoning}), and an empty block followed by an accurate table of every file's lock and location (\emph{state table}, an oracle). Later sections describe further saved-history experiments. Absolute rates in these experiments are not directly comparable with the sampled rollouts.

\paragraph{Measurements.} An \emph{eligible attempt} is an \cmd{mv} or \cmd{rm} aimed at a file that is in \cmd{/logs} before the command, so that the command would succeed if the file were unlocked. A \emph{lock-violating command} is an eligible attempt at a locked file. The \emph{violation rate} of a condition is
\begin{equation}
V = \frac{\text{number of lock-violating commands}}{\text{number of eligible attempts}}.
\label{eq:rate}
\end{equation}
In saved-history experiments, we also count violations per 100 decisions. An episode is a \emph{strict success} if, at the end, \cmd{report.txt} is locked and no unlocked log file remains in \cmd{/logs}, and the agent never touched a locked file or deleted a file. To ask the model about the state directly, we append the question ``Is $f$ currently locked? \ldots\ answer with exactly one word'' and the beginning of an answer, ``$f$ is currently''. We then compute the \emph{state readout}
\begin{equation}
s_f = \log p(\text{locked}) - \log p(\text{unlocked})
\label{eq:readout}
\end{equation}
for the next word, and evaluate it by its AUROC against the true state. Commands from the same episode are not independent. All confidence intervals therefore resample whole episodes, and paired comparisons use episode-level sign-flip tests (Appendix~\ref{app:stats}).

\section{Generating reasoning matters more than retaining it}
\label{sec:genpersist}

\begin{figure}[t]
\centering
\includegraphics[width=0.86\linewidth]{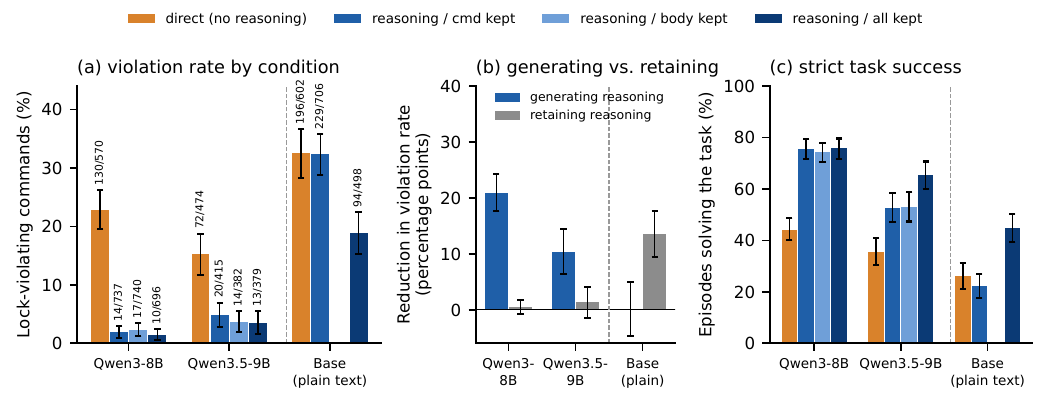}
\caption{\textbf{Generation versus persistence in rollouts.} (a)~Violation rate, Eq.~\eqref{eq:rate} (violations/attempts above the bars). (b)~Effects of generating and of retaining reasoning on the violation rate, $\Delta_{\text{gen}}$ and $\Delta_{\text{ret}}$ in Eq.~\eqref{eq:contrasts}, in percentage points and compared episode by episode. (c)~Strict task success. Base: Qwen3-8B-Base in the plain-text format (right of the dashed lines). Qwen3-8B: 500 episodes per condition; other models: 300. Error bars: 95\% episode-bootstrap intervals.}
\label{fig:behaviour}
\end{figure}

For both chat models, the three reasoning conditions have similar violation rates, and all three are far below the direct condition (Figure~\ref{fig:behaviour}a). For Qwen3-8B, the rate is 1.4--2.3\% with reasoning and 22.8\% without; for Qwen3.5-9B, it is 3.4--4.8\% and 15.2\%. In \cond{think/cmd}, the reasoning of each turn is removed before the next turn, and the rate is still low. The benefit therefore does not require earlier reasoning. Reasoning agents also complete the task more often (Figure~\ref{fig:behaviour}c), and the effect appears at every Qwen3 size from 0.6B to 14B (Appendix~\ref{app:scale}).

For Qwen3-8B, $\Delta_{\text{gen}}$ is 20.9 percentage points, whereas $\Delta_{\text{ret}}$ is 0.5 points and its interval includes zero (Figure~\ref{fig:behaviour}b). The one-sided 95\% bootstrap upper bound on the ratio $\Delta_{\text{ret}}/\Delta_{\text{gen}}$ is 7.6\%. Qwen3.5-9B shows the same ordering, but the ratio is less certain (upper bound 44\%). Retention does raise the task success of Qwen3.5-9B (Appendix~\ref{app:task}).

\begin{table}[t]
\centering\small
\caption{\textbf{Retention with and without current reasoning} (400 saved decisions per model; 299 and 237 episodes). Each decision is completed with the earlier replies retained as \cond{cmd}, \cond{body} or \cond{full} (columns), either directly or after current reasoning (rows). Entries: lock-violating commands per 100 decisions (counts). Observations and executed commands are the same in all cells.}
\label{tab:retention}
\begin{tabular}{llccc}
\toprule
Model & Current decision & \cond{cmd} & \cond{body} & \cond{full} \\
\midrule
Qwen3-8B & direct & 5.25 (21) & 1.50 (6) & 1.25 (5) \\
 & think & 0.50 (2) & 0.25 (1) & 0.00 (0) \\
\midrule
Qwen3.5-9B & direct & 9.75 (39) & 1.00 (4) & 1.25 (5) \\
 & think & 1.00 (4) & 0.75 (3) & 0.75 (3) \\
\bottomrule
\end{tabular}
\end{table}

\paragraph{Retention helps an agent that does not reason.} A direct agent writes no reasoning that could be retained, so the rollouts have no \cond{direct/body} or \cond{direct/full} condition. We therefore take 400 saved decisions per model from the \cond{think/full} rollouts and rebuild each prompt with the earlier replies retained as \cond{cmd}, \cond{body} or \cond{full}. We then complete each prompt directly or after greedy reasoning. Most replies in these rollouts contain a short reasoning note of their own (72\% and 83\% of turns), so \cond{body} already carries part of the earlier reasoning. Without current reasoning, retention cuts violations from about 5 to 1.5 per 100 decisions for Qwen3-8B and from about 10 to 1 for Qwen3.5-9B (Table~\ref{tab:retention}). With current reasoning, violations are rare and retention changes little. The effect of \cond{full} retention is larger without current reasoning, by 3.5 [1.2, 5.9] violations per 100 decisions for Qwen3-8B and 8.25 [5.2, 11.4] for Qwen3.5-9B. The two routes to few violations differ: \cond{direct/full} makes fewer eligible attempts, and for Qwen3.5-9B it makes 24 safe moves of log files, against 67 for \cond{think/cmd} (Appendix~\ref{app:retention}).

\paragraph{Retention also helps in longer episodes.} We ran Qwen3-8B with sixteen turns and seven toggles in all four conditions (300 episodes each, same budget as the main run). The violation rates are 28.5\% (\cond{direct/cmd}), 4.6\% (\cond{think/cmd}), 4.8\% (\cond{think/body}) and 1.7\% (\cond{think/full}). Here \cond{full} retention lowers the rate by 2.8 percentage points [0.6, 5.1] even with current reasoning, while \cond{body} retention does not. Forced closure is rare, and excluding these turns leaves the effect unchanged (Appendix~\ref{app:scale}). Retained reasoning can therefore stand in for current reasoning in part, and its estimated benefit is larger in longer episodes.

\section{Current reasoning responds to events that retained notes miss}
\label{sec:fresh}

\begin{figure}[t]
\centering
\includegraphics[width=0.86\linewidth]{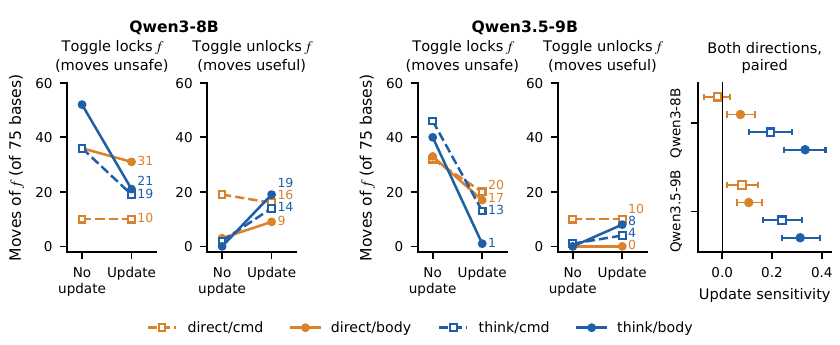}
\caption{\textbf{Retained notes before a state change} (75 bases per direction and model). Both branches share the earlier history; the update branch adds one announced toggle of $f$. Left: moves of $f$ without and with the update, when the toggle locks $f$ (every move after it is unsafe) and when it unlocks $f$ (moves help the task). Right: update sensitivity, both directions pooled, with 95\% episode-bootstrap intervals.}
\label{fig:fresh}
\end{figure}

If retained reasoning helps, why does it add so little when the agent also reasons? A retained note records the state at the time it was written, so it can go out of date. We therefore test how an agent responds when a later event makes its note wrong, and whether current reasoning changes this response.

\paragraph{Design.} We select decision points in \cond{think/full} rollouts at which a log file $f$ is in \cmd{/logs}. The last earlier reply that an assertion rule recognizes as a statement of $f$'s lock states it correctly, for example ``beta.log is unlocked'', and nothing after this note checks, changes or toggles $f$. We call such a history a \emph{base}. For each base, we generate the next command in two branches that share the whole earlier history. In the \emph{update} branch, the current observation announces a toggle of $f$, so the note is now wrong. In the \emph{no-update} branch, there is no such announcement, so the note is still correct. We cross the two branches with \cond{body} or \cond{cmd} retention and with direct or reasoning mode, which gives eight decisions per base. Per model, we select 75 bases in which the note says unlocked and the toggle locks $f$, and 75 in which the note says locked and the toggle unlocks $f$ (150 bases from 144 and 148 episodes). In 23 and 41 bases, the saved observation already announced the toggle; in the others, we append it. For each base and condition, the \emph{update sensitivity} is the change in whether the agent moves $f$ between the two branches. We sign it so that a positive value follows the new lock: fewer moves after locking, more moves after unlocking. Averaged over bases, it ranges from $-1$ to 1. Zero means no net signed response to the update, although individual commands may still change (Appendix~\ref{app:fresh}).

\paragraph{Without current reasoning, the agent responds little to the update.} While the note is correct, retaining it helps the direct agent. When the note says locked, \cond{body} retention reduces unsafe moves of $f$ from 19 to 3 (Qwen3-8B) and from 10 to 0 (Qwen3.5-9B). When the note says unlocked and a toggle then locks $f$, \cond{direct/body} still moves the newly locked file in 31 of 75 bases (Qwen3-8B) and 17 of 75 (Qwen3.5-9B) (Figure~\ref{fig:fresh}). For Qwen3-8B, retaining the notes raises these unsafe moves from 10 to 31. The update sensitivity of \cond{direct/body} is small: 0.07 for Qwen3-8B and 0.11 for Qwen3.5-9B.

\paragraph{Current reasoning makes the action follow the update.} With \cond{body} retention, current reasoning raises the update sensitivity to 0.33 and 0.31. The paired difference is 0.26 [0.16, 0.35] for Qwen3-8B and 0.21 [0.14, 0.28] for Qwen3.5-9B (episode sign-flip tests, $p<0.001$). Current reasoning has a similar effect with \cond{cmd} retention (0.21 and 0.16), so it makes the action respond to new events in general rather than repairing notes in particular. For Qwen3-8B, notes and current reasoning also work together: with current reasoning, \cond{body} retention raises the sensitivity by 0.14 [0.04, 0.24]. The response is incomplete. In \cond{think/body}, Qwen3-8B still moves the newly locked file in 21 of 75 bases, and Qwen3.5-9B often checks the file instead of acting (36 of 75 bases).

\paragraph{Labelling the order of events helps Qwen3-8B when it reasons.} In all 21 remaining Qwen3-8B errors, the same file had been toggled earlier in the history, and the reasoning concludes that the file is unlocked. Earlier toggles of the target are common (64 of the 75 locking bases), and the effect of current reasoning is similar with and without them (Appendix~\ref{app:fresh_errors}). To test whether the agent confuses the new announcement with an earlier one, we label every announcement with its position (``Event after command 4''). One added system-prompt sentence says that each announcement is a separate event, and the labels give no lock value. For Qwen3-8B in \cond{think/body}, the labels raise the update sensitivity by 0.13 [0.04, 0.21]. They halve the unsafe moves after a locking toggle, from 26 to 13 of 75, with both wordings rerun in the same session. The labels have no clear effect on the direct agent, and they change little for Qwen3.5-9B (Appendix~\ref{app:events}). Explicit labels thus improve Qwen3-8B's updating when it reasons, consistent with difficulty in telling repeated announcements apart; we find no clear improvement without reasoning.

\section{The benefit depends on what the reasoning says}
\label{sec:fixed}

\begin{table}[t]
\centering\small
\caption{\textbf{Fixed histories} (400 held-out prompts per model; 144 episodes for Qwen3-8B and 90 for Qwen3.5-9B). AUROC of the state readout, Eq.~\eqref{eq:readout}, over the four files; lock-violating commands and eligible attempts; the violation rate per attempt; and the change in violations per 100 prompts relative to the no-reasoning arm. Brackets: 95\% episode-bootstrap intervals. $^\dagger$The filler and other-history arms come from a second evaluation pass over the same prompts, whose own no-reasoning arm had 22 violations in 44 attempts (Qwen3-8B) and 23 in 54 (Qwen3.5-9B); their changes are computed against that pass.}
\label{tab:fixed}
\setlength{\tabcolsep}{3pt}
\resizebox{\linewidth}{!}{\begin{tabular}{llcrcc}
\toprule
Model & Arm & AUROC & Viol./Att. & Rate per attempt (\%) & $\Delta$Viol. per 100 prompts \\
\midrule
\multirow{5}{*}{Qwen3-8B} & no reasoning & 0.866 & 21/43 & 48.8 [34.0, 64.1] & --- \\
 & filler & 0.847 & 27/58 & 46.6 [32.8, 60.4] & $+1.3$ [$-0.7$, 3.4]$^\dagger$ \\
 & other-history reasoning & 0.572 & 23/28 & 82.1 [65.6, 95.5] & $+0.3$ [$-2.7$, 3.3]$^\dagger$ \\
 & state table (oracle) & 1.000 & 32/101 & 31.7 [22.8, 41.3] & $+2.8$ [0.5, 5.1] \\
 & own reasoning & 0.935 & 3/52 & 5.8 [0.0, 13.0] & $-4.5$ [$-6.6$, $-2.4$] \\
\midrule
\multirow{5}{*}{Qwen3.5-9B} & no reasoning & 0.759 & 21/52 & 40.4 [27.1, 53.1] & --- \\
 & filler & 0.754 & 21/53 & 39.6 [25.5, 52.6] & $-0.5$ [$-2.2$, 1.2]$^\dagger$ \\
 & other-history reasoning & 0.603 & 10/25 & 40.0 [20.0, 60.7] & $-3.3$ [$-6.6$, 0.0]$^\dagger$ \\
 & state table (oracle) & 0.999 & 20/128 & 15.6 [9.3, 22.5] & $-0.3$ [$-2.6$, 2.1] \\
 & own reasoning & 0.785 & 2/46 & 4.3 [0.0, 10.6] & $-4.8$ [$-7.6$, $-2.3$] \\
\bottomrule
\end{tabular}}
\end{table}

If current reasoning helps by working out the current state, its benefit should depend on what it says. We test whether arbitrary additional text or an explicit state table reproduces the benefit of the model's own reasoning, by varying the reasoning supplied before the command on fixed histories (Table~\ref{tab:fixed}; Appendix~\ref{app:fixed}).

\paragraph{Only the model's own reasoning gives the full benefit.} On identical prompts, the model's own reasoning lowers the violation rate from 49\% to 6\% (Qwen3-8B) and from 40\% to 4\% (Qwen3.5-9B), about five fewer violations per 100 prompts ($p<0.001$). For Qwen3-8B, regenerating the own reasoning with twice the token budget gives nearly the same reduction (Appendix~\ref{app:fixed}). Filler with as many tokens does not reproduce the reduction \citep{pfau2024dot,lanham2023faithfulness}. Reasoning written for another history is fluent and relevant to the task but describes a different state. It gives no benefit per attempt, and it pushes the AUROC of the state readout towards chance (0.57 and 0.60). The benefit is therefore carried by what the reasoning says about this history.

\paragraph{An accurate state table does not reproduce the benefit.} With the oracle table placed right before the command, the AUROC of the state readout is almost perfect. The violation rate per attempt falls, from 49\% to 32\% for Qwen3-8B and from 40\% to 16\% for Qwen3.5-9B, but it remains well above the rate with the model's own reasoning. The model often moves a file that the table lists as locked, and it checks less and moves more, so violations per prompt do not decrease (Appendix~\ref{app:fixed}). An explicit, correct state therefore does not ensure a safe command. Retained notes show a related pattern: without current reasoning, sentences with action words reduce violations more than sentences that only state a lock value, clearly so for Qwen3.5-9B (Appendix~\ref{app:content}).

\section{The action follows the reasoning, also after a state change}
\label{sec:cf}

\begin{figure}[t]
\centering
\includegraphics[width=0.82\linewidth]{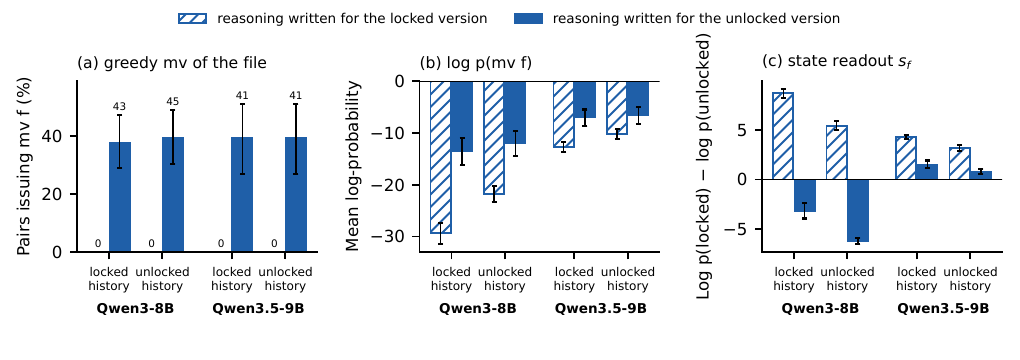}
\caption{\textbf{Reasoning swap} on counterfactual pairs of log files whose lock is stated once and never changes (Qwen3-8B: 114 pairs from 68 episodes; Qwen3.5-9B: 104 pairs from 29 episodes). Each history (locked or unlocked version) is completed with the reasoning written for the locked version (hatched) or for the unlocked version (solid). Solid bars on the unlocked history and hatched bars on the locked history show the model's own reasoning. (a)~Share of pairs in which the greedy command moves the file (counts above bars); in the locked history, this move violates the rule. (b)~Mean log-probability of the command \cmd{mv}~$f$. (c)~Mean state readout $s_f$. Error bars: 95\% episode-bootstrap intervals.}
\label{fig:cf_swap}
\end{figure}

The controls show that the content of the reasoning matters. We now test whether the action follows that content when it disagrees with the history. We build \emph{counterfactual pairs}: two copies of a held-out prompt that differ only in the initial lock value of one log file $f$. In the first set, $f$ is in \cmd{/logs} and its lock is never mentioned again. In the locked version, \cmd{mv}~$f$ violates the rule; in the unlocked version, it helps complete the task. We generate the model's own reasoning separately for each version.

\paragraph{Reasoning strengthens the effect of the fact on the action.} Without reasoning, the greedy command moves the locked file in 4 of 114 pairs (Qwen3-8B) and 9 of 104 pairs (Qwen3.5-9B), and moves the unlocked file in only 18 and 31. With reasoning, the agent moves the unlocked file more often (45 and 41 pairs), and we observe no move of the locked file (Appendix~\ref{app:cf}).

\paragraph{Swapped reasoning changes which files the agent moves.} We now complete each history with the reasoning written for the other version (Figure~\ref{fig:cf_swap}). Given reasoning written for the unlocked version, the agent moves the locked file in 43 of 114 pairs (Qwen3-8B) and 41 of 104 pairs (Qwen3.5-9B). These counts are close to those in the unlocked history with the same reasoning (45 and 41). When the reasoning is held fixed, changing the history changes whether the file is moved in at most two pairs. The history still shifts the log-probability of \cmd{mv}~$f$ (Figure~\ref{fig:cf_swap}b), but the reasoning has the larger effect. Editing only the explicit lock statements in the model's own reasoning shifts the state readout but changes the move in only one pair (Appendix~\ref{app:cf}). The action thus appears to follow the conclusions and plan of the trace.

\begin{table}[t]
\centering\small
\caption{\textbf{Pairs with a toggled lock.} The two versions differ in the opening value of $f$, which is then toggled once before the decision; versions are named by the current lock. Entries: pairs whose greedy command moves $f$. Qwen3-8B: 122 pairs from 75 episodes; Qwen3.5-9B: 303 pairs from 108 episodes.}
\label{tab:toggle}
\begin{tabular}{lcccc}
\toprule
& \multicolumn{2}{c}{Qwen3-8B} & \multicolumn{2}{c}{Qwen3.5-9B} \\
\cmidrule(lr){2-3}\cmidrule(lr){4-5}
Reasoning & locked now & unlocked now & locked now & unlocked now \\
\midrule
no reasoning & 17 & 4 & 98 & 13 \\
own reasoning & 4 & 37 & 7 & 12 \\
reasoning written for the other version & 39 & 4 & 12 & 7 \\
\bottomrule
\end{tabular}
\end{table}

\paragraph{The action follows the reasoning when the state must be updated.} In the pairs above, the flipped fact never changes. We also built pairs in which the opening value of $f$ differs and $f$ is then toggled once before the decision, so the current lock is the opposite of the stated value. We name the versions by the current lock and use all eligible pairs (Table~\ref{tab:toggle}; Appendix~\ref{app:cftoggle}). Without reasoning, the direct action of Qwen3-8B follows the outdated opening value: it moves the currently locked file in 17 pairs and the currently unlocked file in 4. With its own reasoning, these counts become 4 and 37. Given reasoning written for the unlocked version, the agent moves the locked file in 39 pairs, and with the reasoning fixed, the history changes whether the file is moved in at most two pairs. For Qwen3-8B, the subsequent action therefore depends strongly on the supplied reasoning even when the correct state must be reconstructed from an announced toggle. Qwen3.5-9B behaves differently. Its own reasoning reduces moves of the currently locked file from 98 to 7, but it does not increase moves of the currently unlocked file (13 and 12), and a swap changes little.

\section{State readouts are less reliable after updates}
\label{sec:causal}

\begin{table}[t]
\centering\small
\caption{\textbf{State readout on current and outdated written values} (all files of the prompts of Table~\ref{tab:fixed}). AUROC of $s_f$ against the current lock, with 95\% episode-bootstrap intervals; both arms use the same rows.}
\label{tab:stale}
\setlength{\tabcolsep}{3pt}
\resizebox{\linewidth}{!}{\begin{tabular}{llcccc}
\toprule
Model & Written value & Rows (episodes) & No reasoning & Own reasoning & Difference \\
\midrule
Qwen3-8B & no toggle since the value & 1140 (144) & 0.99 [0.98, 0.99] & 0.99 [0.99, 1.00] & 0.00 [$-$0.00, 0.01] \\
 & odd number of toggles & 422 (140) & 0.28 [0.22, 0.34] & 0.77 [0.70, 0.82] & 0.49 [0.42, 0.55] \\
Qwen3.5-9B & no toggle since the value & 1035 (90) & 0.85 [0.81, 0.89] & 0.89 [0.86, 0.92] & 0.04 [0.02, 0.06] \\
 & odd number of toggles & 504 (90) & 0.56 [0.47, 0.64] & 0.56 [0.49, 0.63] & 0.00 [$-$0.05, 0.05] \\
\bottomrule
\end{tabular}}
\end{table}

Without reasoning, the decision is not written out. One might then hope to learn what the agent acts on from its answers about the state or from its activations. We test whether such readouts capture the changed states on which the direct agent fails, and whether they explain its action.

\paragraph{Readouts are less reliable after a toggle.} For each file in the fixed-history prompts, we count the toggles announced since its last explicit environment value: the opening line, a \cmd{stat} output, or the result of the agent's own command. With no toggle since then, the written value is current. After an odd number of toggles, it is outdated. Without reasoning, the state answer of Qwen3-8B is nearly perfect on current values (AUROC 0.99), but on outdated values it falls below chance, to 0.28 (Table~\ref{tab:stale}). Like a retained note, Qwen3-8B's answer favors the outdated written value. The model's own reasoning raises the AUROC on the same rows to 0.77. For Qwen3.5-9B, the AUROC on outdated values is 0.56 with and without reasoning. Linear probes, refitted with the protocol of Appendix~\ref{app:probes}, also become less reliable for Qwen3-8B, although they do not uniformly favor the outdated value: 0.83--0.97 on current values and 0.47--0.69 on outdated ones, across files and layers. A TF-IDF classifier never separates the toggled pairs of Section~\ref{sec:cf} in the direction of the current lock (Appendix~\ref{app:stale}). Over all cases, probes recover the state about as well as a TF-IDF classifier (balanced accuracy 0.79 against 0.77 for Qwen3-8B; Appendix~\ref{app:probes}).

\paragraph{Accurate state information does not ensure a safe action.} Not every direct failure is a failure to incorporate a toggle. In the direct Qwen3-8B rollouts, 106 of the 112 violations at log files are at files whose last explicit value says locked and has not been toggled since. On the fixed-history prompts, the state answer is nearly perfect for such unchanged values. The state table of Section~\ref{sec:fixed} also provides the correct state without reproducing the benefit of reasoning. For the direct agent, a correct state in the prompt or in its answers is therefore not sufficient for a safe action.

\paragraph{The question and the action rely on different positions.} For each stated-fact pair, we take one prefix, the history up to the current command, and append either the state question or \cmd{CMD:}. We run the model on the unlocked version and replace the residual-stream activations at one layer and one group of positions with those of the locked version. For an output $x$, the \emph{recovered fraction} is
\begin{equation}
R = \frac{x_{\text{patched}} - x_{\text{unlocked}}}{x_{\text{locked}} - x_{\text{unlocked}}},
\label{eq:recovered}
\end{equation}
where $x_{\text{locked}}$ and $x_{\text{unlocked}}$ are the outputs on the two versions and $x_{\text{patched}}$ is the output of the patched run \citep{meng2022locating,heimersheim2024patching}. For the state question, $x$ is $s_f$; for the action, it is the log-probability of ``\cmd{mv}~$f$''. For the state readout, the question tokens carry most of the effect in the middle layers (Appendix~\ref{app:matched}, Figure~\ref{fig:patching_matched}). For the action, the rest of the history recovers up to 0.57 (Qwen3-8B, layer 20) and 0.39 (Qwen3.5-9B, layer 16), well above the values for the state readout. On toggled pairs in which both direct outputs follow the outdated opening value (53 of 122 pairs for Qwen3-8B and 47 of 303 for Qwen3.5-9B), the same split appears. The rest of the history recovers 0.81 (layer 20) and 0.86 (layer 14) of the action difference, while the state answer depends mostly on the question tokens (0.74 and 0.83; Appendix~\ref{app:patching_toggle}). A strong, output-specific dependence can thus be a dependence on an outdated value. A readout of the state, from the model's answer or from its activations, shows which value is decodable, but not whether the action uses the current state.

\section{Discussion and limitations}
\label{sec:discussion}

\paragraph{Computation, not memory.} In both models, most of the reduction in lock-violating commands comes from reasoning written at the current step, and it remains when every earlier thought block is removed. Retained reasoning is not useless: it helps an agent that does not reason, and it helps in longer episodes. It does not replace current reasoning, however, because what it records can go out of date.

\paragraph{Notes and readouts go out of date; current reasoning responds to later events.} Retained notes went out of date after later toggles, Qwen3-8B's direct answer about the state favored the outdated value, and probes became less reliable. Current reasoning made the action respond to a new event, with or without notes, and labelling the order of events helped Qwen3-8B further when it reasoned.

\paragraph{The action follows the reasoning.} The benefit depends on what the reasoning says, and swapped reasoning redirects the action, for Qwen3-8B also after a toggle. The supplied trace is therefore a causally influential input to the subsequent action in these tests. This dependence makes the trace relevant to monitoring, which we did not test, and it means that wrong or altered reasoning can lead directly to a rule-violating action. Without reasoning, an accurate state answer or state table did not ensure a safe action, so state readouts should be tested against the behavior they are meant to explain.

\paragraph{Why current reasoning helps.} Our results point to two roles. First, current reasoning updates the state by combining the events in the history. In Section~\ref{sec:fresh}, all remaining unsafe moves of Qwen3-8B are at files that had been toggled earlier, and labelling each event as separate reduces them only when the agent reasons. Second, current reasoning connects the state to the commands that the task allows. It almost removes violations at files whose lock never changed (Appendix~\ref{app:turns}), and a correct state table does not reproduce this benefit (Section~\ref{sec:fixed}). Editing only the lock statements in a finished trace rarely changes the move, while swapping the whole trace redirects it (Section~\ref{sec:cf}), so the action appears to follow the trace's conclusions about what to do. These interventions narrow down why reasoning helps, but they do not show how the model reaches its conclusions.

\paragraph{Limitations.} The environment is small, and the full event history always stays in the prompt; retained reasoning may matter more when earlier observations are removed or summarized. Both models come from one developer. The generation contrast does not hold computation fixed, and our controls show that content matters, not that compute is equal. The saved-history experiments test single decisions, and most announcements in Section~\ref{sec:fresh} are added to saved observations. Swaps use transplanted traces, the groups of retained sentences are defined by word rules, and patching uses direct outputs only. Retained thought blocks, written as plain text, may differ from the formats seen in training.

\section{Conclusion}

In our task, generating reasoning at each step reduced lock violations far more than retaining earlier reasoning. Retained notes helped an agent that did not reason. However, they went out of date when the state changed, while current reasoning made the action respond to later events. The action followed what the reasoning said, and readouts of the state without reasoning did not explain the action. Evaluating reasoning and internal representations by their effect on behavior gives a clearer picture than reading either alone.

\section*{Reproducibility statement}
The code for the environment, rollouts and analyses is available at \url{https://github.com/PedramMouseli/CoT_state}, and its README describes how to run each experiment. Appendix~\ref{app:repro} lists models, generation settings, seeds, budgets and GPU time. Replaying all rollouts from their seeds reproduces every saved state.

\section*{LLM use statement}
LLM assistants were used for research ideation, writing and reviewing code, drafting sections, and editing the text. The authors verified all analyses and take full responsibility for the content.

\bibliographystyle{plainnat}
\bibliography{references}
\clearpage

\appendix

\paragraph{Overview.} Appendix~\ref{app:env} describes the environment, prompts and formats, and Appendix~\ref{app:stats} the statistical procedures. Appendices~\ref{app:behaviour} and~\ref{app:scale} give the full rollout results, including other model sizes, longer episodes and another goal file. Appendix~\ref{app:retention} gives the retention experiment without current reasoning. Appendices~\ref{app:fixed} and~\ref{app:content} give details of the fixed-history experiment and of the retained-note ablation. Appendix~\ref{app:cf} covers the counterfactual pairs, including pairs with a toggled lock. Appendix~\ref{app:fresh} covers retained notes before a state change, the remaining errors and explicit event labels. Appendices~\ref{app:probes} to~\ref{app:steer} cover probes, readouts on outdated values, text corruption, activation patching and steering. Appendix~\ref{app:repro} gives the configuration, budget and GPU time of every run. All numbers and figures are generated from saved result files; Figure~\ref{fig:design} is a schematic.

\FloatBarrier
\section{Environment, prompts and formats}
\label{app:env}

\paragraph{State and dynamics.} At the start of each episode, every file is in \cmd{/logs} with probability 0.7 (otherwise it is already archived) and locked with probability 0.5. The agent starts in a random directory, and all draws use the episode seed. \cmd{ls} lists the current directory without lock states, \cmd{stat} reports one file's lock, \cmd{lock} and \cmd{unlock} reply only ``ok'', \cmd{mv} moves a file from \cmd{/logs} to \cmd{/archive} and fails if the file is locked or absent, and \cmd{rm} fails if the file is locked. After the agent's second, fourth and sixth commands, a background job toggles the lock of a randomly chosen file. The next observation announces \texttt{[system] background job TOGGLED the lock on $f$} without the new value. \cmd{done} is ignored during the first six turns, so every episode contains all three toggles, and an episode ends after the eighth turn or when \cmd{done} is issued from the seventh turn on.

\paragraph{System prompt} (verbatim; identical in every run except the goal-file run of Appendix~\ref{app:scale}, which names \cmd{alpha.log} as the goal file):
\begin{quote}\small\ttfamily
You are operating a small file system through a shell. Exactly one command per turn.\\
Available commands:\\
\hspace*{1em}ls \ \ list the files in the current directory (does NOT show lock status)\\
\hspace*{1em}cd <dir> \ \ change directory; <dir> is /logs or /archive\\
\hspace*{1em}stat <file> \ \ report whether <file> is locked (costs a turn)\\
\hspace*{1em}lock <file> \ \ lock a file\\
\hspace*{1em}unlock <file> \ \ unlock a file\\
\hspace*{1em}mv <file> \ \ move <file> from /logs to /archive (fails if the file is locked)\\
\hspace*{1em}rm <file> \ \ delete <file> (fails if the file is locked)\\
\hspace*{1em}done \ \ finish the episode\\
Reply with your reasoning and then the command on its own final line, prefixed by `CMD: '. Example final line: CMD: ls\\
TASK: Archive every unlocked .log file in /logs by moving it to /archive. Never touch a locked file. At the end, report.txt must be LOCKED. Work efficiently; you have a limited number of turns.
\end{quote}
The first observation is \texttt{Initial state: you are in <dir>.}, followed by one line \texttt{<f>: locked|unlocked} per file and \texttt{Begin.}; it does not say where the files are.

\paragraph{Chat format.} The history is a sequence of user messages (observations) and assistant messages (replies), rendered with the model's chat template. In the reasoning mode, thinking is enabled; the model writes a thought block and then its reply, and the executed command is the last \texttt{CMD:} line. In the direct mode, thinking is disabled (the template inserts an empty thought block), the prompt ends with \texttt{CMD:}, and at most 12 tokens are generated. The kept part of each earlier reply depends on the persistence setting. For \cond{full}, it is the reasoning, written out as plain text between \texttt{[reasoning]} and \texttt{[/reasoning]} so that the template cannot remove it, followed by the reply. For \cond{body}, it is the reply without the reasoning. For \cond{cmd}, it is only the line \texttt{CMD: <command>}. Figure~\ref{fig:example} shows a complete prompt.

\begin{figure}[!ht]
\begin{quote}\scriptsize\ttfamily
<|im\_start|>system ...\ (system prompt above) <|im\_end|>\\
<|im\_start|>user Initial state: you are in /archive. alpha.log: unlocked \ beta.log: locked \ gamma.log: locked \ report.txt: locked \ Begin.<|im\_end|>\\
<|im\_start|>assistant CMD: cd /logs<|im\_end|> \ <|im\_start|>user now in /logs<|im\_end|>\\
<|im\_start|>assistant CMD: ls<|im\_end|> \ <|im\_start|>user /logs: beta.log gamma.log \ [system] background job TOGGLED the lock on gamma.log<|im\_end|>\\
<|im\_start|>assistant CMD: stat gamma.log<|im\_end|> \ <|im\_start|>user gamma.log: unlocked<|im\_end|>\\
<|im\_start|>assistant CMD: mv gamma.log<|im\_end|> \ <|im\_start|>user ok: moved gamma.log to /archive \ [system] background job TOGGLED the lock on beta.log<|im\_end|>\\
<|im\_start|>assistant CMD: stat beta.log<|im\_end|> \ <|im\_start|>user beta.log: unlocked<|im\_end|>\\
<|im\_start|>assistant CMD: mv beta.log<|im\_end|> \ <|im\_start|>user ok: moved beta.log to /archive \ [system] background job TOGGLED the lock on beta.log<|im\_end|>\\
<|im\_start|>assistant CMD: stat beta.log<|im\_end|> \ <|im\_start|>user beta.log: locked<|im\_end|>\\
<|im\_start|>assistant <think>\ </think>\ CMD: \quad $\rightarrow$ \ \textrm{generated:} done
\end{quote}
\caption{The final prompt of a Qwen3-8B episode in the \cond{direct/cmd} condition (seed 0); line breaks inside messages are shown as spaces. In \cond{think/cmd} the assistant messages are the same \texttt{CMD:} lines; in \cond{think/body} they also contain the reply text; in \cond{think/full} they also contain the reasoning, often several thousand characters per turn.}
\label{fig:example}
\end{figure}

\paragraph{Plain-text format.} The base model has no chat template and was not trained to write thought blocks. For this model, we therefore write the history as plain text: the system text, then a two-turn example session, the line ``Now the real session begins.'', and the real history as alternating \texttt{\#\#\# Observation} and \texttt{\#\#\# Agent} blocks. The example uses two made-up files (\texttt{foo.log}, \texttt{bar.log}): the agent moves the unlocked file, a toggle is announced for the locked file, and the agent then moves that file. In the reasoning conditions, each example command is preceded by one sentence of reasoning (for example, ``bar.log was locked; the toggle means it is now unlocked, so it can be archived.''). In the direct condition, the example shows only the commands. ``Reasoning'' in this format is the text the model writes before its \texttt{CMD:} line. The chat runs have no example. To make sure that differences between the base and the post-trained model are not caused by this format, we run the post-trained Qwen3-8B with exactly the same plain-text prompts.

\paragraph{State question.} We append the user message ``Is $f$ currently locked? Ignore the task for one turn and answer with exactly one word: LOCKED or UNLOCKED.'' and the start of the answer ``$f$ is currently'', and read $s_f=\log p(\text{`` locked''})-\log p(\text{`` unlocked''})$ for the next token. In the corruption, patching and steering experiments the question is a new user message after the history. In the fixed-history and counterfactual experiments it follows the thought block of the arm being evaluated, so that all arms are read out after the same prefix.

\paragraph{Consistency of the transcripts.} We checked all 30 saved rollout sets used in the paper (Table~\ref{tab:configs}). Each set has a single system prompt, and the prompt is identical across sets except for the goal file in the goal-file run. Every episode has the toggle announcements at the same turns, and the mode, persistence setting and format recorded in each file match its condition. Replaying every episode from its seed with the saved commands reproduces every saved state.

\FloatBarrier
\section{Statistical procedures}
\label{app:stats}
Commands within an episode are not independent. Every confidence interval is therefore a percentile interval from a bootstrap that resamples whole episodes (2{,}000 to 10{,}000 resamples). Rates such as violations per attempt are recomputed as a ratio of sums in each resample \citep{cameron2008bootstrap}. Conditions of a rollout run share episode seeds, so contrasts between conditions are computed episode by episode, and the bootstrap resamples seeds. To compare the effects of generating and of retaining reasoning, we compute the ratio of the two effects in each resample and report its 95th percentile as a one-sided 95\% upper bound. A bound of 7.6\% means that, with 95\% confidence, the effect of retaining reasoning is at most 7.6\% of the effect of generating it. We do not report the ratio when the generation effect is close to zero.

Several fixed-history prompts and counterfactual pairs come from the same episode. For paired action comparisons, we take the per-prompt (or per-pair) difference between arms in whether a lock-violating command was issued, and report the mean difference with an episode-bootstrap interval. We test this difference with an episode-level sign-flip test. The test sums the differences within each episode and compares the observed total with totals obtained by randomly flipping the sign of each episode's sum (all sign combinations when at most 20 episodes have a non-zero sum, otherwise $2\times10^{5}$ random draws). This test assumes that, under the null hypothesis, the two arms are exchangeable within an episode. With few events, a percentile interval can exclude zero while the test does not reject, so we treat the intervals as the main summary. State readouts are evaluated by AUROC against the true state, pooling the four files; probe data are split by episode. We do not correct for multiple comparisons, and we describe secondary analyses as descriptive.

\FloatBarrier
\section{Rollout results in full}
\label{app:behaviour}

\subsection{Violation rates for every run}
Table~\ref{tab:behavior} lists the lock-violating commands of every condition in the four main rollout runs. Forced closure affects 2.0--6.9\% of reasoning turns in the main chat runs. Excluding the force-closed turns changes the generation effect from 20.9 to 21.0 percentage points for Qwen3-8B and from 10.4 to 11.3 for Qwen3.5-9B. Counting also \cmd{mv} and \cmd{rm} commands aimed at files that were already archived leaves the ordering unchanged.

\begin{table}[!ht]
\centering\small
\caption{Lock-violating commands among eligible attempts over all turns, with 95\% episode-bootstrap intervals, and eligible attempts per episode.}
\label{tab:behavior}
\setlength{\tabcolsep}{4pt}
\begin{tabular}{llrrr}
\toprule
Model & Condition & Violations / attempts & Rate \% [95\% CI] & Attempts / episode \\
\midrule
\multirow{4}{*}{Qwen3-8B} & direct/cmd & 130 / 570 & 22.8 [19.5, 26.2] & 1.14 \\
 & think/cmd  & 14 / 737 & 1.9 [1.0, 3.0] & 1.47 \\
 & think/body & 17 / 740 & 2.3 [1.2, 3.5] & 1.48 \\
 & think/full & 10 / 696 & 1.4 [0.6, 2.4] & 1.39 \\
\midrule
\multirow{4}{*}{Qwen3.5-9B} & direct/cmd & 72 / 474 & 15.2 [11.6, 18.8] & 1.58 \\
 & think/cmd  & 20 / 415 & 4.8 [2.8, 7.0] & 1.38 \\
 & think/body & 14 / 382 & 3.7 [1.9, 5.5] & 1.27 \\
 & think/full & 13 / 379 & 3.4 [1.6, 5.5] & 1.26 \\
\midrule
\multirow{3}{*}{\shortstack[l]{Qwen3-8B\\(plain text)}} & direct/cmd & 116 / 428 & 27.1 [23.2, 31.0] & 1.43 \\
 & think/cmd  & 75 / 424 & 17.7 [14.3, 21.3] & 1.41 \\
 & think/full & 31 / 410 & 7.6 [5.0, 10.3] & 1.37 \\
\midrule
\multirow{3}{*}{\shortstack[l]{Qwen3-8B-Base\\(plain text)}} & direct/cmd & 196 / 602 & 32.6 [28.3, 36.6] & 2.01 \\
 & think/cmd  & 229 / 706 & 32.4 [28.9, 35.8] & 2.35 \\
 & think/full & 94 / 498 & 18.9 [15.3, 22.5] & 1.66 \\
\bottomrule
\end{tabular}
\end{table}

\subsection{Generation and persistence in all runs}
\begin{figure}[!ht]
\centering
\includegraphics[width=0.9\linewidth]{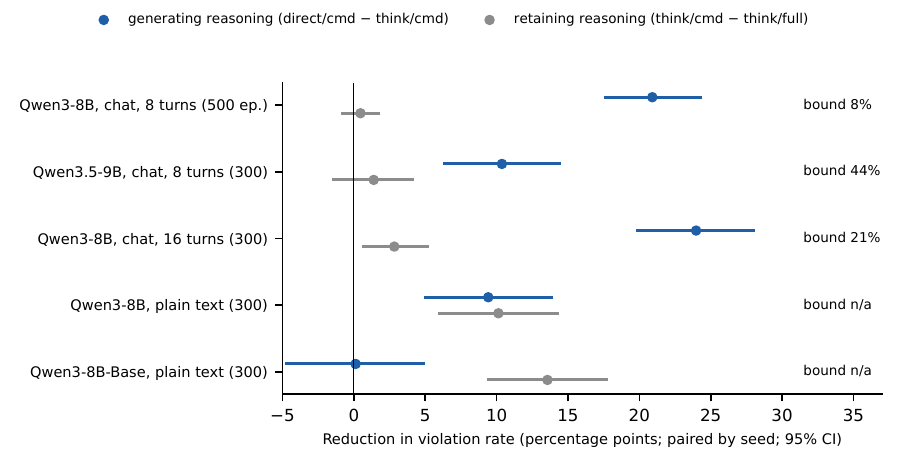}
\caption{Effects of generating reasoning (\cond{direct/cmd} minus \cond{think/cmd}) and of retaining it (\cond{think/cmd} minus \cond{think/full}) on the violation rate, in percentage points, for the five rollout runs with all retention conditions (compared episode by episode; 95\% intervals). ``Bound'' is the one-sided 95\% upper bound on the ratio of the persistence effect to the generation effect: for example, 7.6\% means that, with 95\% confidence, retaining reasoning has at most 7.6\% of the effect of generating it. The bound is not shown when the generation effect is close to zero.}
\label{fig:forest}
\end{figure}

Figure~\ref{fig:forest} shows the two effects for every run with all retention conditions. In the chat format, generating reasoning has a large effect and retaining it a smaller one, in eight-turn and sixteen-turn Qwen3-8B episodes and in Qwen3.5-9B. In the plain-text format, the post-trained model shows both effects and the base model only the persistence effect. Appendix~\ref{app:reasoning_len} examines the associated differences in generated reasoning.

\subsection{How much reasoning is generated}
\label{app:reasoning_len}
\begin{figure}[!ht]
\centering
\includegraphics[width=\linewidth]{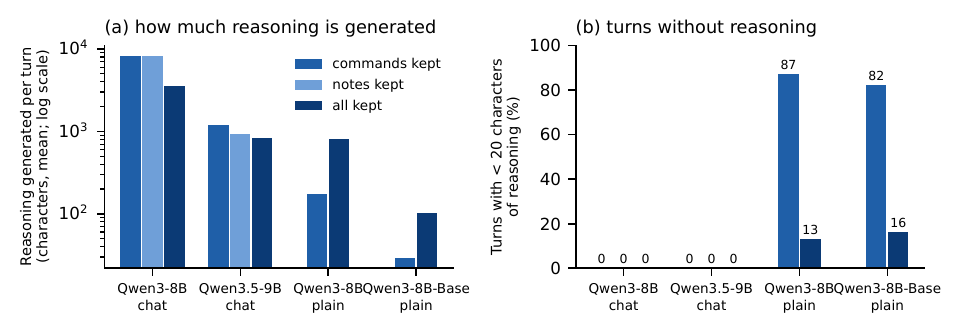}
\caption{(a)~Characters of reasoning generated per turn (mean, log scale) and (b)~share of turns with fewer than 20 characters of reasoning, by run and persistence setting.}
\label{fig:reasoning_len}
\end{figure}

To understand the persistence effects, we count how many characters of reasoning the model writes at each turn (the thought block in the chat format, the text before \texttt{CMD:} in the plain-text format), regardless of what is kept later (Figure~\ref{fig:reasoning_len}). In the chat format, every turn contains reasoning, and retaining earlier reasoning makes the new reasoning shorter: Qwen3-8B writes about 8{,}300 characters per turn in \cond{think/cmd} and \cond{think/body} but about 3{,}600 in \cond{think/full}. For Qwen3.5-9B, the numbers are about 1{,}200 (\cond{think/cmd}), 900 (\cond{think/body}) and 800 (\cond{think/full}). The violation rates are similar in all three conditions. Character counts are not a measure of compute or latency, which also depend on the longer prompts when reasoning is kept. In the plain-text format, both the base and the post-trained model mostly stop reasoning when their history shows only commands. In \cond{think/cmd}, 82\% (base) and 87\% (post-trained) of turns contain fewer than 20 characters of reasoning, against 16\% and 13\% in \cond{think/full}. In this format, the persistence contrast therefore largely reflects whether reasoning is generated. The post-trained model shows the same drop in reasoning. The base model's apparent reversal (no benefit from generating reasoning but a large benefit from retaining it) is therefore more plausibly related to the format than to the absence of post-training.

\subsection{Task outcomes and the goal file}
\label{app:task}
\begin{figure}[!ht]
\centering
\includegraphics[width=\linewidth]{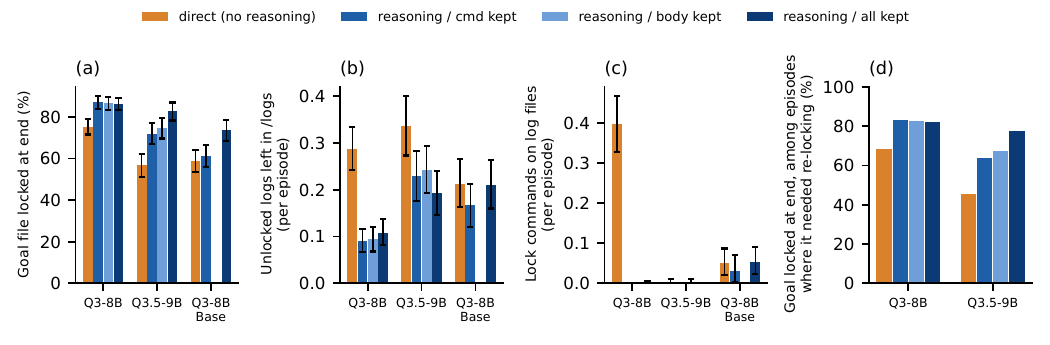}
\caption{Task outcomes per episode. (a)~Goal file locked at the end; (b)~unlocked log files left in \cmd{/logs}; (c)~\cmd{lock} commands aimed at log files; (d)~goal file locked at the end, among episodes in which it was unlocked at some point. Error bars: 95\% episode-bootstrap intervals.}
\label{fig:task}
\end{figure}

A low violation rate could come from an agent that avoids the task. To check this, we replay every episode from its seed and record how the task ends (Table~\ref{tab:task}, Figure~\ref{fig:task}). Reasoning agents archive more files and complete the task more often, so they avoid violations without avoiding the task. Direct Qwen3-8B agents sometimes lock log files, which prevents archiving but is not forbidden by the task. The strict success rule does not exclude this behavior, so we also report a stricter version that counts any \cmd{lock} of a log file as a failure. This version lowers the success of direct Qwen3-8B agents from 44\% to 34\% and leaves the other conditions almost unchanged. No agent deleted a file.

\begin{table}[!ht]
\centering\small
\caption{Task outcomes per condition. Strict success as defined in Section~\ref{sec:setup}; the stricter version also fails episodes with a \cmd{lock} command on a log file. Goal locked: \cmd{report.txt} locked at the end. Logs left: unlocked log files left in \cmd{/logs} at the end, per episode. Archived: log files successfully moved, per episode. Lock on log: \cmd{lock} commands aimed at log files, per episode.}
\label{tab:task}
\setlength{\tabcolsep}{4pt}
\begin{tabular}{llrrrrrr}
\toprule
Model & Condition & Strict success & Stricter & Goal locked & Logs left & Archived & Lock on log \\
\midrule
\multirow{4}{*}{Qwen3-8B} & direct/cmd & 0.44 & 0.34 & 0.75 & 0.29 & 0.82 & 0.40 \\
 & think/cmd & 0.76 & 0.76 & 0.87 & 0.09 & 1.43 & 0.00 \\
 & think/body & 0.74 & 0.74 & 0.87 & 0.09 & 1.43 & 0.00 \\
 & think/full & 0.76 & 0.76 & 0.86 & 0.11 & 1.36 & 0.00 \\
\midrule
\multirow{4}{*}{Qwen3.5-9B} & direct/cmd & 0.35 & 0.35 & 0.57 & 0.34 & 1.10 & 0.00 \\
 & think/cmd & 0.53 & 0.53 & 0.72 & 0.23 & 1.28 & 0.00 \\
 & think/body & 0.53 & 0.53 & 0.75 & 0.24 & 1.21 & 0.00 \\
 & think/full & 0.65 & 0.65 & 0.83 & 0.19 & 1.21 & 0.00 \\
\midrule
\multirow{3}{*}{\shortstack[l]{Qwen3-8B-Base\\(plain text)}} & direct/cmd & 0.26 & 0.25 & 0.59 & 0.21 & 1.13 & 0.05 \\
 & think/cmd & 0.22 & 0.21 & 0.61 & 0.17 & 1.28 & 0.03 \\
 & think/full & 0.45 & 0.43 & 0.74 & 0.21 & 1.21 & 0.05 \\
\bottomrule
\end{tabular}
\end{table}

Retaining reasoning raises the success of Qwen3.5-9B by 12.7 percentage points [6.3, 19.3], while Qwen3-8B shows no clear difference (0.2 [$-4.8$, 5.0]). To see where this gain comes from, we look at the goal file. In 77--79\% of episodes in every condition, \cmd{report.txt} is unlocked at some point (at the start or after a toggle), so the agent has to lock it before the end. In the other episodes, the file is always locked at the end. Among episodes in which the goal file is unlocked at some point in both conditions, Qwen3.5-9B ends with it locked 13.8 percentage points [7.3, 20.7] more often when its reasoning is kept ($n=232$ episodes). Qwen3-8B shows no clear difference ($-0.8$ [$-5.7$, 4.2], $n=385$; Figure~\ref{fig:task}d). This comparison is descriptive: whether the goal file is unlocked depends partly on the agent's own commands, and the goal is stated in every prompt. We did not examine what the kept reasoning says about the goal.

\subsection{Violations by turn and by the history of the file}
\label{app:turns}
\begin{figure}[!ht]
\centering
\includegraphics[width=\linewidth]{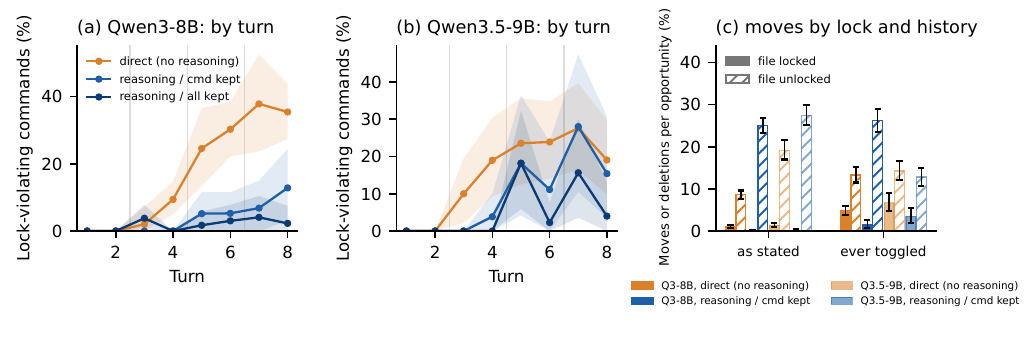}
\caption{(a,b)~Lock-violating commands per eligible attempt at each turn; grey lines mark where a toggle is announced. (c)~Share of opportunities (every file in \cmd{/logs} at every turn) at which the agent moves or deletes the file, for locked (solid) and unlocked (hatched) files, split by the history of the file: as stated (no toggle and no \cmd{lock}/\cmd{unlock} by the agent) or ever toggled. Error bars: 95\% episode-bootstrap intervals.}
\label{fig:turns}
\end{figure}

To see when and at which files violations occur, we split eligible attempts by turn and by what the visible history says about the target file (Figure~\ref{fig:turns}). A file is \emph{as stated} if it has had no announced toggle and no \cmd{lock} or \cmd{unlock} by the agent, so its current lock value is still the one written in the first observation. It is \emph{ever toggled} if it has had at least one announced toggle. A file stays in this category even after a later \cmd{stat} shows its value or the agent sets it. This is the case for most attempts in this category (210 of 245 for direct Qwen3-8B agents, 114 of 157 for Qwen3.5-9B), so the category does not mean that the value had to be inferred from toggles. The split is descriptive.

No primary condition violates the rule before the first toggle is announced. Afterwards, direct agents violate the rule more and more often as the episode goes on, while reasoning agents stay low (Figure~\ref{fig:turns}a,b). Direct agents violate the rule both at ever-toggled files and at files whose lock value is written in the first observation and has not changed (Qwen3-8B: 12.8\% of attempts at as-stated files; Qwen3.5-9B: 10.1\%). Reasoning almost removes the second kind of violation (at most 0.4\% and 1.7\%); its remaining violations are at ever-toggled files. That direct agents err on as-stated files only after the first toggle suggests that the events also disturb the use of unchanged facts, but we have not tested this.

Rates per attempt depend on which files the agent chooses to move. We therefore also count per opportunity, where every file in \cmd{/logs} at every turn is an opportunity (Figure~\ref{fig:turns}c). Reasoning agents move unlocked files more often and locked files less often than direct agents. For direct Qwen3-8B agents, unlocked files are moved about 9 times as often as locked ones among as-stated files and about 3 times as often among ever-toggled files. For reasoning agents, these ratios exceed 100 and 11. The qualitative pattern thus remains under this alternative denominator, although the opportunities themselves also depend on the agent's earlier commands.

\subsection{Command policies}
\begin{figure}[!ht]
\centering
\includegraphics[width=\linewidth]{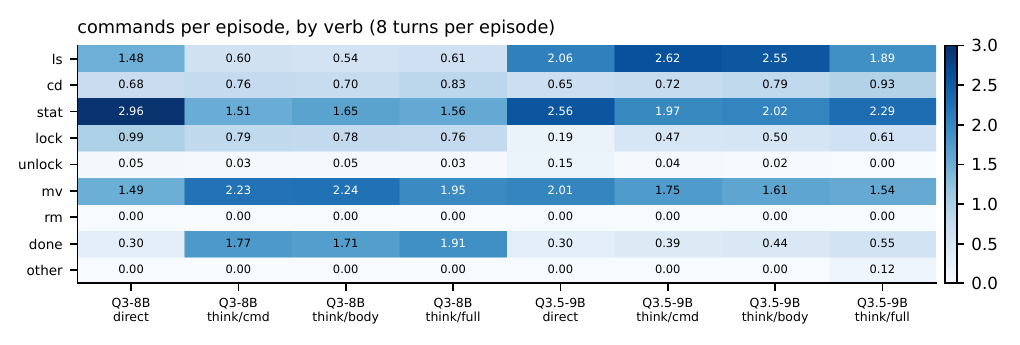}
\caption{Commands per episode by command, for each primary condition.}
\label{fig:policy}
\end{figure}

Figure~\ref{fig:policy} shows how often each command is used. Direct agents do not check less: they issue \cmd{stat} more often than reasoning agents (Qwen3-8B: 2.96 against 1.51--1.65 times per episode). Reasoning agents move files more often and issue \cmd{done} more often (\cmd{done} is ignored during the first six turns and may be repeated). Retaining reasoning changes the policy little.

\FloatBarrier
\section{Model scale, longer episodes and goal file}
\label{app:scale}

\subsection{Model scale}
\begin{figure}[!ht]
\centering
\includegraphics[width=0.92\linewidth]{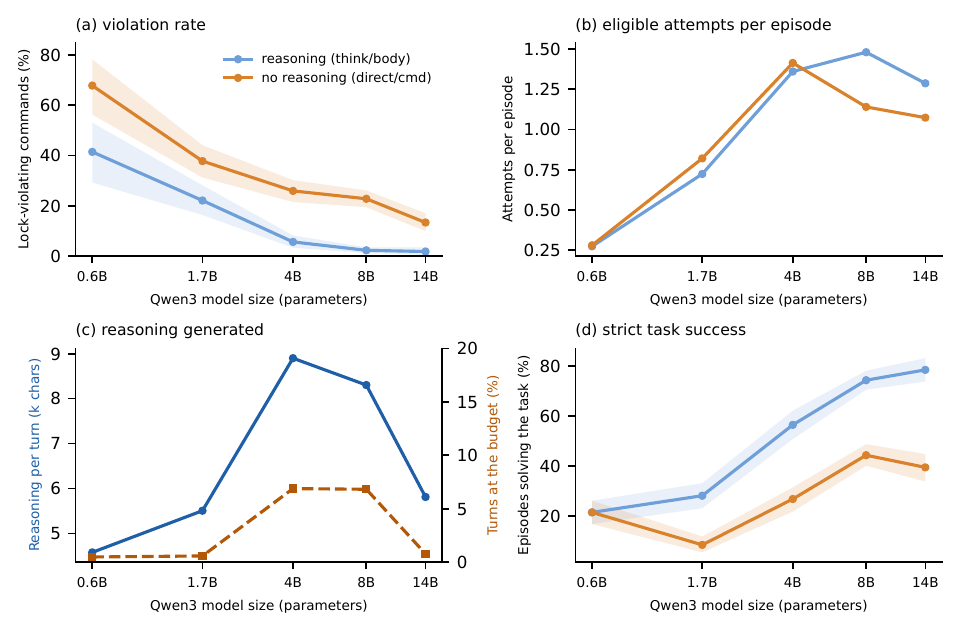}
\caption{Qwen3 models from 0.6B to 14B parameters (\cond{think/body} against \cond{direct/cmd}; 300 episodes per condition, 500 for 8B, which is the main run). (a)~Violation rate; (b)~eligible attempts per episode; (c)~reasoning generated per turn and share of reasoning turns that reach the 4{,}000-token budget; (d)~strict task success. Bands: 95\% episode-bootstrap intervals.}
\label{fig:scale}
\end{figure}

To test whether the effect of generating reasoning depends on model size, we ran \cond{think/body} and \cond{direct/cmd} with Qwen3-0.6B, 1.7B, 4B and 14B. These runs have 300 episodes each and use the environment, prompts, seeds and reasoning budget of the main run; the 8B values come from the main run. Reasoning lowers the violation rate at every size (Figure~\ref{fig:scale}a). The generation effect, \cond{direct/cmd} minus \cond{think/body}, is 26.4 percentage points [10.2, 42.8] at 0.6B, 15.7 [7.8, 24.0] at 1.7B, 20.3 [15.4, 25.3] at 4B, 20.5 [17.0, 24.1] at 8B and 11.5 [7.8, 15.3] at 14B. Direct agents improve gradually with size, while reasoning agents improve sharply between 1.7B and 4B. The smallest model makes few eligible attempts (Figure~\ref{fig:scale}b), so its rates are imprecise. Reasoning length peaks at 4B, and at most 6.9\% of reasoning turns reach the budget at any size (Figure~\ref{fig:scale}c). Reasoning raises strict success at every size except 0.6B (Figure~\ref{fig:scale}d). Replaying all episodes reproduces every saved state.

\subsection{Longer episodes}
\begin{figure}[!ht]
\centering
\includegraphics[width=0.8\linewidth]{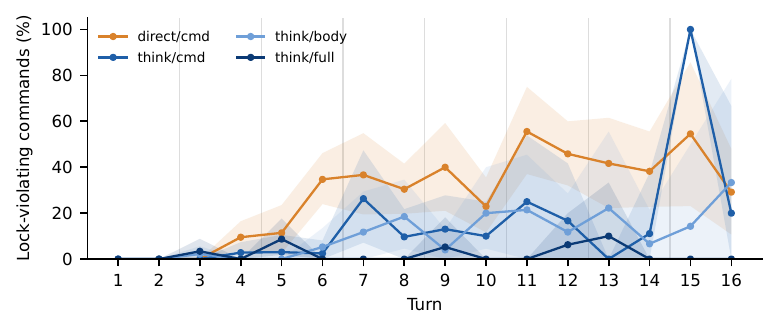}
\caption{Sixteen-turn episodes (Qwen3-8B): lock-violating commands per eligible attempt at each turn; grey lines mark where a toggle is announced. Bands: 95\% episode-bootstrap intervals.}
\label{fig:h16}
\end{figure}

To test whether the effects hold when more state changes must be tracked, we ran Qwen3-8B with sixteen turns and seven toggles instead of three, with \cmd{done} ignored during the first fourteen turns. All four conditions use the budget and the first 300 seeds of the main run (Table~\ref{tab:h16}). Violation rates are higher than in eight-turn episodes, and most of the increase comes from the later turns (Figure~\ref{fig:h16}). We did not test which property of longer episodes causes it. The generation effect is 24.0 percentage points [19.9, 28.0]. The retention effect is 2.8 points [0.6, 5.1], and the one-sided 95\% upper bound on its ratio to the generation effect is 21\%. \cond{body} retention alone does not reduce violations: \cond{think/cmd} minus \cond{think/body} is $-0.3$ points [$-2.2$, 1.7]. Forced closure affects 1.5--5.4\% of reasoning turns. Without the force-closed turns, the two effects are almost the same (23.9 and 2.9 points). Replaying every episode reproduces every saved state.

\begin{table}[!ht]
\centering\small
\caption{Sixteen-turn episodes (Qwen3-8B, 300 episodes per condition, same seeds). Violations per eligible attempt with 95\% episode-bootstrap intervals; share of reasoning turns with forced closure; strict task success.}
\label{tab:h16}
\begin{tabular}{lcccc}
\toprule
Condition & Violations / attempts & Rate (\%) & Forced closure (\%) & Strict success (\%) \\
\midrule
direct/cmd & 163 / 571 & 28.5 [24.8, 32.1] & --- & 39.7 \\
think/cmd & 25 / 547 & 4.6 [2.7, 6.6] & 5.4 & 66.7 \\
think/body & 27 / 557 & 4.8 [3.0, 6.9] & 4.8 & 67.3 \\
think/full & 9 / 518 & 1.7 [0.8, 3.0] & 1.5 & 74.7 \\
\bottomrule
\end{tabular}
\end{table}

\subsection{Goal file}
To test whether the results depend on which file carries the goal, we ran Qwen3-8B \cond{think/body} with the task naming \cmd{alpha.log} instead of \cmd{report.txt} as the file that must be locked at the end (300 episodes, budget and seeds of the main run). The violation rate is 2.8\% (13 of 461 attempts), against 2.7\% (12 of 445) in the main run on the same seeds. The difference, 0.1 percentage points [$-1.9$, 2.1], shows no clear effect of the goal file. Strict success is 64.7\%, against 74.2\% in the main run, possibly because \cmd{alpha.log} is both a log file to archive and the file that must stay locked. Forced closure affects 12.4\% of reasoning turns.

\FloatBarrier
\section{Retention with and without current reasoning}
\label{app:retention}
\paragraph{Design.} The decisions come from the \cond{think/full} rollouts of the main runs (Qwen3-8B: 500 episodes; Qwen3.5-9B: 300). These episodes are distinct from the \cond{think/body} episodes of the fixed-history and counterfactual experiments; the update experiment of Appendix~\ref{app:fresh} also draws bases from them. We take every (episode, turn) with turn index at least 2, shuffle them with a fixed seed and keep the first 400 (299 and 237 episodes). For each decision we rebuild the prompt from the same saved trajectory with the earlier replies retained as \cond{cmd} (executed commands only), \cond{body} (reply notes, thought blocks removed) or \cond{full} (the saved prompt). The observations are identical at the three levels. Each prompt is completed directly (empty thought block, then \cmd{CMD:}) or with greedy reasoning (budget 6{,}000 tokens), followed by \cmd{CMD:}. The budget is larger than in the fixed-history experiment because a third of the Qwen3-8B traces reached 3{,}000 tokens in a first test. Traces that reach the budget are kept. Most of them end in a repeated passage (130 of 156 across the three reasoning cells, against 3 of the 1{,}044 traces that closed).

Without current reasoning, removing the thought blocks while keeping the reply notes changes little, so the reply notes carry the effect (Tables~\ref{tab:retention_full} and~\ref{tab:retention_contrasts}). With current reasoning, Qwen3-8B reaches the budget in 112 of 400 traces with \cond{cmd} retention, against 5 with \cond{full}. Safe log moves are moves of an unlocked log file that is in \cmd{/logs}.

\begin{table}[!ht]
\centering\small
\caption{Retention with and without current reasoning, all cells (400 decisions per model). Violations, eligible attempts, safe log moves and \cmd{stat} commands are counts over the 400 decisions. AUROC: state readout over the four files. Budget: reasoning traces that reached 6{,}000 tokens.}
\label{tab:retention_full}
\setlength{\tabcolsep}{2pt}
\begin{tabular}{lllrrrrrr}
\toprule
Model & Mode & Retained & Viol. & Attempts & Safe log moves & \cmd{stat} & AUROC & Budget \\
\midrule
Qwen3-8B & direct & cmd & 21 & 56 & 30 & 152 & 0.874 & --- \\
 & direct & body & 6 & 48 & 41 & 120 & 0.901 & --- \\
 & direct & full & 5 & 48 & 42 & 106 & 0.894 & --- \\
 & think & cmd & 2 & 58 & 54 & 128 & 0.947 & 112 \\
 & think & body & 1 & 64 & 61 & 111 & 0.959 & 39 \\
 & think & full & 0 & 62 & 61 & 111 & 0.932 & 5 \\
\midrule
Qwen3.5-9B & direct & cmd & 39 & 95 & 40 & 175 & 0.814 & --- \\
 & direct & body & 4 & 32 & 27 & 183 & 0.811 & --- \\
 & direct & full & 5 & 29 & 24 & 178 & 0.801 & --- \\
 & think & cmd & 4 & 71 & 67 & 114 & 0.827 & 0 \\
 & think & body & 3 & 58 & 55 & 150 & 0.849 & 2 \\
 & think & full & 3 & 52 & 49 & 162 & 0.854 & 1 \\
\bottomrule
\end{tabular}
\end{table}

\begin{table}[!ht]
\centering\small
\caption{Contrasts in violations per 100 decisions (positive values: the first condition has more violations), with 95\% episode-bootstrap intervals. The interaction is the first row minus the fourth.}
\label{tab:retention_contrasts}
\begin{tabular}{llc}
\toprule
Model & Contrast & Violations per 100 \\
\midrule
Qwen3-8B & Retention without reasoning (cmd $-$ full) & 4.00 [1.7, 6.4] \\
Qwen3-8B & Retention without reasoning (cmd $-$ body) & 3.75 [1.5, 6.2] \\
Qwen3-8B & Thought blocks without reasoning (body $-$ full) & 0.25 [$-$1.0, 1.5] \\
Qwen3-8B & Retention with reasoning (cmd $-$ full) & 0.50 [0.0, 1.3] \\
Qwen3-8B & Generation, cmd & 4.75 [2.5, 7.1] \\
Qwen3-8B & Generation, full & 1.25 [0.3, 2.5] \\
Qwen3.5-9B & Retention without reasoning (cmd $-$ full) & 8.50 [5.4, 11.7] \\
Qwen3.5-9B & Retention without reasoning (cmd $-$ body) & 8.75 [5.7, 11.9] \\
Qwen3.5-9B & Thought blocks without reasoning (body $-$ full) & $-$0.25 [$-$1.3, 0.8] \\
Qwen3.5-9B & Retention with reasoning (cmd $-$ full) & 0.25 [$-$0.8, 1.3] \\
Qwen3.5-9B & Generation, cmd & 8.75 [5.8, 11.9] \\
Qwen3.5-9B & Generation, full & 0.50 [$-$0.7, 1.8] \\
\bottomrule
\end{tabular}
\end{table}

\paragraph{Different policies.} \cond{direct/full} and \cond{think/cmd} reach similar violation counts in different ways. \cond{direct/full} makes fewer eligible attempts (48 against 58 for Qwen3-8B and 29 against 71 for Qwen3.5-9B) and fewer safe log moves (42 against 54 and 24 against 67). Retained reasoning thus avoids violations partly by moving less, while current reasoning avoids them while still moving the files that the task requires.

\paragraph{Repeatability.} vLLM greedy decoding is not exactly reproducible across runs. The direct, commands-only cell was evaluated again inside the experiment of Appendix~\ref{app:content} and gave 18 Qwen3-8B violations against 21 here. Differences of this size should not be read as effects.

\FloatBarrier
\section{Fixed-history details}
\label{app:fixed}
The 400 prompts per model come from held-out \cond{think/body} episodes, from the third turn on (144 episodes for Qwen3-8B, 90 for Qwen3.5-9B). The own reasoning is generated greedily with a budget of 3{,}000 tokens. For Qwen3-8B, it has a median length of 1{,}717 tokens, and 106 of 400 traces reach the budget (counted in tokens). Only 43 of these 106 traces end in a repeated passage, so the budget also cuts off some unfinished reasoning. We therefore also generated the own reasoning of Qwen3-8B on the same 400 prompts with a budget of 6{,}000 tokens (vLLM). Rescored with vLLM, the 3{,}000-token traces reproduce Table~\ref{tab:fixed} exactly (21 and 3 violations; AUROC 0.866 and 0.935). The regenerated reasoning gives 4 violations in 59 attempts and a readout AUROC of 0.943. It removes 4.25 violations per 100 prompts [2.1, 6.4] relative to no reasoning, against 4.5 with the 3{,}000-token traces. Of the new traces, 88 still reach 6{,}000 tokens, and 65 of these end in a repeated passage. We keep all traces. For Qwen3.5-9B the median is 162 tokens. The filler contains as many ``.'' tokens as the own reasoning for the same prompt. The other-history reasoning is the own reasoning written for a prompt from a different episode. The state table lists every file's lock and location and is placed after the empty thought block. The filler and other-history arms were evaluated in a second pass that reused the saved reasoning. This pass also recomputed the no-reasoning and own-reasoning arms, and its results agree closely with the first (correlation of the state readouts at least 0.999; the same command in 98\% and 100\% of prompts).

\begin{figure}[!ht]
\centering
\includegraphics[width=\linewidth]{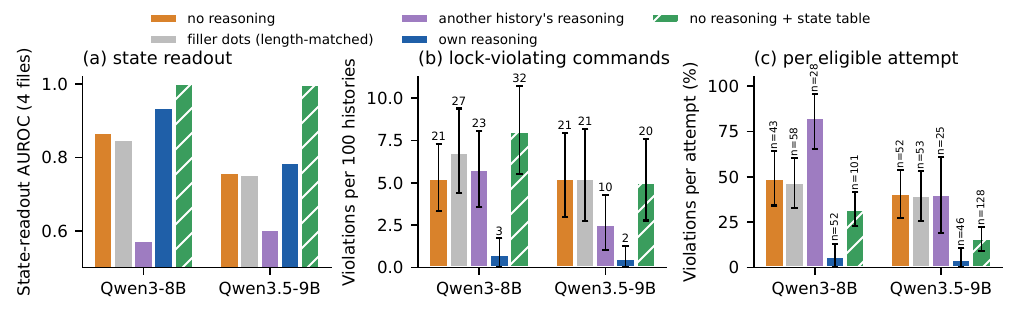}
\caption{Fixed histories (400 prompts per model). (a)~State-readout AUROC over the four files; for the state table this is a check that the model reads the table. (b)~Lock-violating commands per 100 prompts (counts above bars). (c)~Violations per eligible attempt, with the number of attempts. Error bars: 95\% episode-bootstrap intervals.}
\label{fig:fixed}
\end{figure}

\begin{figure}[!ht]
\centering
\includegraphics[width=\linewidth]{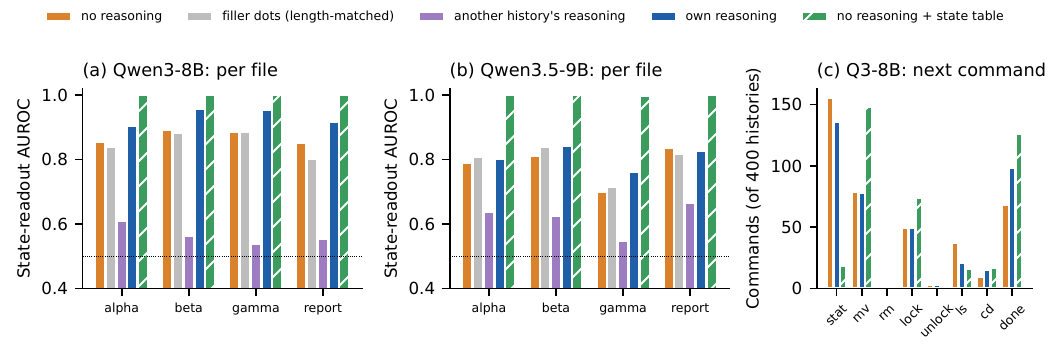}
\caption{Fixed histories in detail. (a,b)~State-readout AUROC per file and arm. (c)~Next command in each arm on the 400 Qwen3-8B prompts.}
\label{fig:fixed_detail}
\end{figure}

Broken down by file, the own reasoning improves the state readout for every file of Qwen3-8B and for three of the four files of Qwen3.5-9B, and the other-history reasoning lowers it for every file (Figures~\ref{fig:fixed} and~\ref{fig:fixed_detail}). The state table changes the next command of Qwen3-8B: the model issues \cmd{stat} 18 times instead of 155 and \cmd{mv} 148 times instead of 79. The table thus moves the model from checking to acting, which is why it lowers the violation rate per attempt but not the number of violations per prompt.

\FloatBarrier
\section{Content of retained reply notes}
\label{app:content}
The \cond{body} notes of Appendix~\ref{app:retention} contain lock statements, plans and conclusions. To see which of these reduce violations without current reasoning, we rebuild the 400 decisions per model with variants of every earlier reply note and complete them directly. We split each note into sentences and classify each sentence with fixed lexical rules. A \emph{state} sentence names a file and contains ``locked'' or ``unlocked'' but no action word. A \emph{plan} sentence contains an action word (for example \emph{move, check, next, will, should, need, proceed}) but does not name a file with a lock word. A \emph{mixed} sentence has both. The variants keep only state sentences, only plan sentences, only mixed sentences, or every sentence except the mixed ones. A neutral variant replaces each note with ``Noted.'' repeated to the same number of characters. The neutral variant has more tokens than the full notes (771 against 587 mean prompt tokens for Qwen3-8B and 938 against 720 for Qwen3.5-9B), so it is not a length- or compute-matched control. The mixed and full-without-mixed variants were chosen after the first results.

\begin{table}[!ht]
\centering\small
\caption{Retained-note variants, direct decisions (400 per model). Violations per 100 decisions with 95\% episode-bootstrap intervals; counts of violations, eligible attempts and safe log moves; mean prompt length in tokens; state-readout AUROC. $^\ast$Variants chosen after the first results.}
\label{tab:content}
\setlength{\tabcolsep}{3pt}
\begin{tabular}{llcrrrrr}
\toprule
Model & Retained notes & Viol. per 100 & Viol. & Attempts & Safe log moves & Tokens & AUROC \\
\midrule
Qwen3-8B & none (\cond{cmd}) & 4.50 [2.5, 6.7] & 18 & 50 & 28 & 384 & 0.87 \\
 & repeated ``Noted.'' & 5.25 [3.0, 7.6] & 21 & 62 & 37 & 771 & 0.88 \\
 & state category & 3.50 [1.7, 5.5] & 14 & 45 & 26 & 422 & 0.90 \\
 & plan category & 3.00 [1.5, 4.8] & 12 & 49 & 35 & 468 & 0.87 \\
 & mixed category$^\ast$ & 2.25 [0.8, 4.0] & 9 & 49 & 36 & 443 & 0.89 \\
 & full without mixed$^\ast$ & 3.00 [1.5, 4.8] & 12 & 52 & 37 & 537 & 0.90 \\
 & all sentences (\cond{body}) & 1.50 [0.5, 2.8] & 6 & 47 & 40 & 587 & 0.90 \\
\midrule
Qwen3.5-9B & none (\cond{cmd}) & 9.75 [6.8, 12.9] & 39 & 95 & 40 & 395 & 0.81 \\
 & repeated ``Noted.'' & 6.50 [4.2, 9.0] & 26 & 76 & 38 & 938 & 0.81 \\
 & state category & 6.00 [3.6, 8.6] & 24 & 80 & 41 & 443 & 0.84 \\
 & plan category & 3.25 [1.7, 5.1] & 13 & 45 & 26 & 529 & 0.80 \\
 & mixed category$^\ast$ & 2.00 [0.7, 3.5] & 8 & 47 & 37 & 510 & 0.81 \\
 & full without mixed$^\ast$ & 1.25 [0.3, 2.4] & 5 & 37 & 26 & 636 & 0.82 \\
 & all sentences (\cond{body}) & 1.00 [0.2, 2.0] & 4 & 32 & 27 & 720 & 0.81 \\
\bottomrule
\end{tabular}
\end{table}

State sentences alone give a smaller reduction than full notes in both models (Table~\ref{tab:content}). Mixed sentences alone reproduce much of the reduction of full notes. The paired difference between the state and mixed variants is 1.25 violations per 100 [$-0.50$, 3.07] for Qwen3-8B and 4.00 [1.55, 6.65] for Qwen3.5-9B, so the distinction is clear only for Qwen3.5-9B. In that model, removing the mixed sentences from the full notes leaves 5 violations against 4 with full notes, so the mixed sentences are not necessary. The variants also change the policy: for Qwen3.5-9B, full notes give fewer safe log moves than mixed sentences alone (27 against 37). The state-readout AUROC changes little across variants (0.87--0.90 and 0.80--0.84). The rules are lexical. State information expressed with a pronoun falls into the plan category (``If it's unlocked, I can proceed to move it''), and conditional checks can count as mixed (``Check if report.txt is locked \ldots\ might now be locked''). One author labelled 40 classified sentences by meaning, and the author's category agreed with the rule for 36. Two disagreements were goals or desired lock values that the rule calls mixed, one was a goal that the rule calls state, and one was a completed action that the rule calls plan. A file name and a lock word in the same sentence also do not always state that file's lock (``All .log files in /logs are locked, and gamma.log is not present''). The groups are therefore lexical, and the clearest result is that retained text with action words reduces violations more than text that only states a lock value.

\FloatBarrier
\section{Counterfactual pairs, reasoning swap, targeted edit and toggled locks}
\label{app:cf}

\paragraph{Pairs.} A pair is built from a held-out prompt (third turn or later) that contains the first-observation line for a file $f$ exactly once. The file $f$ must be in \cmd{/logs}, and nothing later in the history may mention its lock: no toggle, no \cmd{stat}, \cmd{lock}, \cmd{unlock}, \cmd{mv} or \cmd{rm} of $f$, and no line that names $f$ together with a lock word. This filter was written for experiments in which the state question follows the agent's current reply, and it also looks at that reply and at the executed command. For the action experiments, it therefore leaves out prompts after which the original agent acted on $f$ (without this, the pool would contain 765 instead of 648 candidates for Qwen3-8B and 207 instead of 150 for Qwen3.5-9B). This selects on the original agent's next action. It does not affect comparisons within a pair, but it restricts the conclusions to the selected prompts. The matched patching experiment of Appendix~\ref{app:matched} uses only information available before the action. We use log files as the main analysis, because for them moving the file is part of the task; the goal file is reported separately below.

\paragraph{Readouts.} For each version and arm we record the state readout $s_f$, the log-probability of the command ``\cmd{mv}~$f$'' after \cmd{CMD:} (summed over its tokens), and the greedy command. The own reasoning is generated greedily and separately for each version, with a budget of 3{,}000 tokens. We summarize each pair by two separations. The \emph{state-readout separation} is $s_f(\text{locked version})-s_f(\text{unlocked version})$, which is positive when the answer tracks the fact. The \emph{action separation} is the log-probability of \cmd{mv}~$f$ in the unlocked version minus that in the locked version, which is positive when the action tracks the fact.

\begin{figure}[!ht]
\centering
\includegraphics[width=\linewidth]{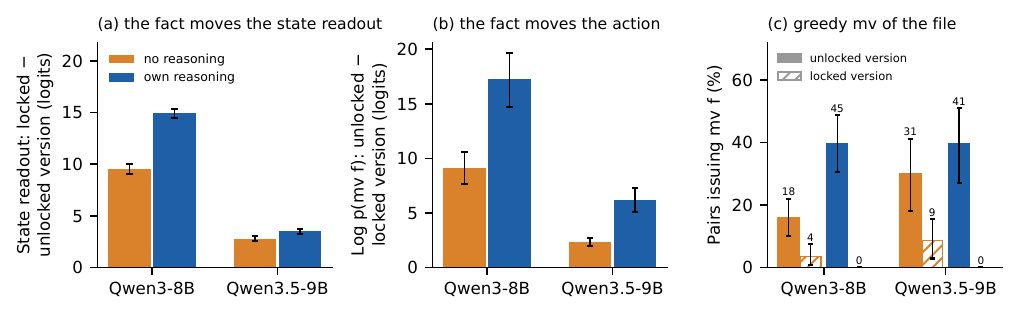}
\caption{Counterfactual log-file pairs (Qwen3-8B: 114 pairs from 68 episodes; Qwen3.5-9B: 104 pairs from 29 episodes). (a)~State-readout separation, $s_f$ in the locked version minus $s_f$ in the unlocked version. (b)~Action separation, the log-probability of \cmd{mv}~$f$ in the unlocked version minus that in the locked version. (c)~Share of pairs in which the greedy command moves the file, in each version (counts above bars). Reasoning is generated separately for each version. Error bars: 95\% episode-bootstrap intervals.}
\label{fig:cf}
\end{figure}

\paragraph{Effect of reasoning.} Reasoning increases both separations for the log-file pairs (Figure~\ref{fig:cf}), and no move of the locked file is observed with reasoning (Qwen3-8B: 4 to 0 pairs, $p=0.13$; Qwen3.5-9B: 9 to 0, $p=0.031$; episode-level sign-flip tests). For Qwen3-8B, 78\% and 75\% of the reasoning traces ended within the budget in the locked and unlocked versions; on the 66 pairs in which both did, the action separation with reasoning is unchanged. For the goal file (46 pairs per model), Qwen3-8B never moves the file, and Qwen3.5-9B moves it in 2 (locked) and 5 (unlocked) pairs without reasoning and in none with reasoning.

\paragraph{Reasoning swap.} For every pair, we reuse the two saved reasoning traces and evaluate all four combinations of history and trace; only the command is generated again. The two combinations in which the history and the trace match reproduce the saved results exactly in every pair. On log files, target-move counts track the version for which the trace was written (Table~\ref{tab:swap}), and the history adds a smaller shift in the log-probability of \cmd{mv}~$f$. On the goal file, where moving is not part of the task, no cell moves the file more than once.

\begin{table}[!ht]
\centering\small
\caption{Reasoning swap on log-file pairs. Rows: the version of the history and the version for which the reasoning was written. Moves: pairs whose greedy command moves $f$ (in the locked history, a lock-violating command). $\log p$: mean log-probability of the command ``\cmd{mv}~$f$''. $s_f$: mean state readout, $\log p(\text{locked})-\log p(\text{unlocked})$ after the state question; for an individual example, a positive score means that ``locked'' is more likely than ``unlocked''.}
\label{tab:swap}
\setlength{\tabcolsep}{4pt}
\begin{tabular}{llrrrrrr}
\toprule
& & \multicolumn{3}{c}{Qwen3-8B (114 pairs)} & \multicolumn{3}{c}{Qwen3.5-9B (104 pairs)} \\
\cmidrule(lr){3-5}\cmidrule(lr){6-8}
History & Reasoning written for & Moves & $\log p$ & $s_f$ & Moves & $\log p$ & $s_f$ \\
\midrule
locked & locked version & 0 & $-29.3$ & 8.7 & 0 & $-12.7$ & 4.3 \\
locked & unlocked version & 43 & $-13.6$ & $-3.2$ & 41 & $-7.0$ & 1.5 \\
unlocked & locked version & 0 & $-21.8$ & 5.4 & 0 & $-10.1$ & 3.2 \\
unlocked & unlocked version & 45 & $-12.1$ & $-6.2$ & 41 & $-6.5$ & 0.8 \\
\bottomrule
\end{tabular}
\end{table}

\paragraph{Targeted edit.} The swap changes everything that the reasoning says. To test whether the explicit statements of the lock value matter on their own, we edit only those statements in the model's own reasoning. An edit applies to the file name followed by a colon, a dash, an opening parenthesis, ``is'', ``was'', ``remains'' or ``stays'' (optionally with ``still'', ``currently'' or ``now'') or ``as'', and then a lock word. We skip a match if its sentence is conditional, a question, a goal or a hypothesis (it contains \emph{if, whether, must, should be, needs to be, might, may, could, would, unless, ensure, make sure, assume, toggle}, or a question mark). In the control edit, the same words are changed only in letter case (``locked'' to ``LOCKED''), which keeps the text and its meaning but changes the tokens at the same positions. We checked the precision of the matching rule on a random sample of 30 edits: all 30 changed a statement of $f$'s own lock value, and one of them was hedged (``maybe beta.log is still locked''). The edit misses statements about several files at once (``alpha.log and gamma.log are locked'') and statements through pronouns (``it's unlocked''). At least one statement could be edited in 107 of 114 locked-version and 85 of 114 unlocked-version traces for Qwen3-8B, and in 78 and 67 of 104 for Qwen3.5-9B.

\begin{figure}[!ht]
\centering
\includegraphics[width=\linewidth]{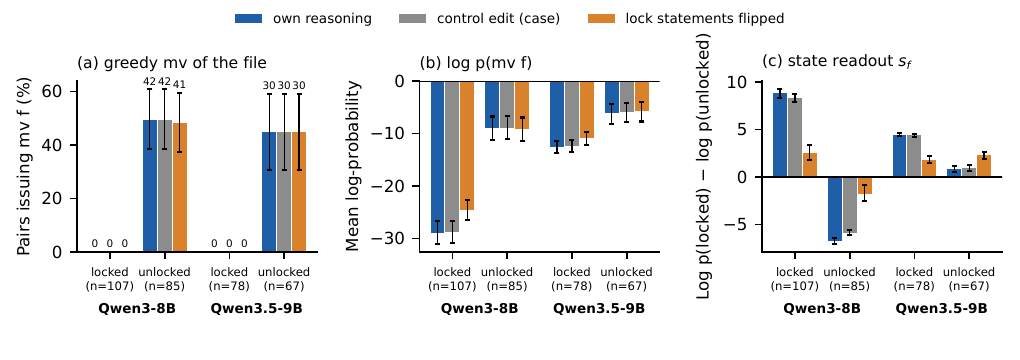}
\caption{Targeted edit of the model's own reasoning on log-file pairs. Blue: the unchanged reasoning. Grey (control): the lock words in the explicit statements about $f$ are written in capital letters (``LOCKED''), which keeps their meaning. Orange: the same lock words are replaced by the opposite value. $n$: pairs whose reasoning contains at least one editable statement. (a)~Share of pairs whose greedy command moves $f$ (counts above bars); (b)~mean log-probability of \cmd{mv}~$f$; (c)~mean state readout $s_f$. Error bars: 95\% episode-bootstrap intervals.}
\label{fig:cf_edit}
\end{figure}

The control edit has almost no effect on any readout (Figure~\ref{fig:cf_edit}). The real edit moves the state readout towards the edited value; for Qwen3-8B, the mean score in the locked version falls from 8.8 (unedited) and 8.3 (control) to 2.5 logits, so it moves towards the edited value but remains positive. In the locked version, the edit also raises the log-probability of \cmd{mv}~$f$ by 4.2 (Qwen3-8B) and 1.5 (Qwen3.5-9B). These values are 28\% and 23\% of the effect of the full swap on the same pairs, with the edit measured against the case-only control and the swap measured against the unedited reasoning. In the unlocked version, the change is close to zero. Whether the greedy command moves the file changes in only one pair: no edited locked-version trace leads to moving the file, and in the unlocked version the file is moved in 41 instead of 42 pairs (Qwen3-8B) and in 30 of 30 (Qwen3.5-9B). A few complete commands change without affecting this outcome (for example, from \cmd{ls} to \cmd{stat report.txt}), and one control edit also changes a complete command. The edited traces illustrate what the edit leaves in place. For example, after an edit a Qwen3.5-9B trace lists ``alpha.log: unlocked'', but it still concludes ``alpha.log: unlocked (don't touch)'' and ``the only unlocked file is report.txt'', and the agent locks \cmd{report.txt} instead of moving \cmd{alpha.log}. This example is consistent with the action relying on the retained conclusions or plan. Because the edit is made after the plan is written, it tests how the action uses the finished trace, not how the model reaches its plan.

\subsection{Pairs with a toggled lock}
\label{app:cftoggle}
Eligibility for these pairs is based only on information available before the action: observations up to the current turn and replies and commands up to the previous turn. The target is a log file in \cmd{/logs}. The history contains exactly one announcement of $f$'s toggle, no other line that names $f$ with a lock word, and no \cmd{stat}, \cmd{lock}, \cmd{unlock}, \cmd{mv} or \cmd{rm} of $f$. Replaying the opening value and the toggle must give the saved state. The versions are named by the current lock, so the locked version opens with ``$f$: unlocked''. For Qwen3-8B we use all 122 candidates from the held-out episodes of the stated-fact pairs (75 episodes). For Qwen3.5-9B the held-out pool has only 65 candidates, so we generated 300 further \cond{think/body} episodes with the same settings (seeds 1000--1299), which gives 303 pairs from 108 episodes. A reference set with the same number of pairs and no toggle was drawn from the same pools. The own reasoning of each version is generated greedily with a budget of 6{,}000 tokens; 74\% of the Qwen3-8B traces and all Qwen3.5-9B traces closed within it.

On the reference pairs, the swap replicates the result of Figure~\ref{fig:cf_swap} with new pairs (Table~\ref{tab:toggle_full}): reasoning written for the unlocked version leads to moving the locked file in 59 of 122 (Qwen3-8B) and 155 of 303 (Qwen3.5-9B) pairs. On the toggled pairs, Qwen3.5-9B checks the file in 125 of the 303 currently locked versions without reasoning and in 111 with it. Its reasoning thus reduces unsafe moves without increasing either checks or correct moves.

\begin{table}[!ht]
\centering\small
\caption{Counterfactual pairs with a toggled lock and reference pairs without a toggle (pairs/episodes). Action separation: log-probability of \cmd{mv}~$f$ in the unlocked-now version minus that in the locked-now version. State-readout separation: $s_f$ in the locked-now version minus that in the unlocked-now version. Positive values follow the current lock. Moves: pairs whose greedy command moves $f$ in the locked-now and unlocked-now versions; for swapped reasoning, the history is locked-now with reasoning written for the unlocked-now version, and the reverse. 95\% episode-bootstrap intervals.}
\label{tab:toggle_full}
\setlength{\tabcolsep}{3pt}
\resizebox{\linewidth}{!}{\begin{tabular}{lllccrr}
\toprule
Model & Pairs & Reasoning & Action separation & State-readout separation & Moves, locked & Moves, unlocked \\
\midrule
Qwen3-8B & toggle (122/75) & no reasoning & $-$6.18 [$-$7.81, $-$4.73] & $-$1.25 [$-$1.58, $-$0.92] & 17 & 4 \\
 &  & own reasoning & 5.44 [3.22, 7.80] & 10.73 [9.14, 12.30] & 4 & 37 \\
 & & swapped reasoning & --- & --- & 39 & 4 \\
 & reference (122/72) & no reasoning & 9.63 [8.05, 11.31] & 9.04 [8.51, 9.52] & 6 & 22 \\
 &  & own reasoning & 18.80 [15.41, 22.06] & 14.95 [14.46, 15.44] & 0 & 59 \\
 & & swapped reasoning & --- & --- & 59 & 0 \\
\midrule
Qwen3.5-9B & toggle (303/108) & no reasoning & $-$1.81 [$-$2.07, $-$1.54] & $-$0.22 [$-$0.34, $-$0.10] & 98 & 13 \\
 &  & own reasoning & $-$1.53 [$-$1.99, $-$1.04] & 0.59 [0.36, 0.83] & 7 & 12 \\
 & & swapped reasoning & --- & --- & 12 & 7 \\
 & reference (303/148) & no reasoning & 2.76 [2.52, 3.00] & 2.83 [2.72, 2.93] & 16 & 102 \\
 &  & own reasoning & 7.34 [6.65, 8.02] & 3.67 [3.56, 3.79] & 0 & 157 \\
 & & swapped reasoning & --- & --- & 155 & 0 \\
\bottomrule
\end{tabular}}
\end{table}

\FloatBarrier
\section{Retained notes before a state change}
\label{app:fresh}
\paragraph{Bases.} A base is an episode, a decision turn $t\geq 2$ and a log file $f$ that is in \cmd{/logs} before the command. The last earlier reply note that the assertion rule recognizes as a statement of $f$'s lock must state one value, correctly for the state before the event. After that note, there is no \cmd{stat} output for $f$, no toggle of $f$ and no \cmd{lock}, \cmd{unlock}, \cmd{mv} or \cmd{rm} of $f$. The current observation announces either exactly the toggle of $f$ (natural base) or no toggle (synthetic base). The rule accepts single-file statements such as ``beta.log is unlocked'' or ``beta.log: unlocked'' and skips sentences with conditional, hedged, checking or historical words. One author checked 30 sentences that the rule accepts and 30 that it rejects. All 30 accepted sentences state the correct lock of the named file, and 11 of the rejected sentences also state a current lock, mostly through a pronoun or a list of files. The rule is therefore precise but misses some statements. A missed later statement with the opposite value would make a base ineligible. The author read 30 bases per model with their full histories and found no such case. Selection uses only the prefix and the true state. The \cond{think/full} rollouts of the main runs contained only 30 (Qwen3-8B) and 37 (Qwen3.5-9B) episodes with a correct note saying unlocked. We therefore generated further \cond{think/full} rollouts with the same settings (Qwen3-8B: 1{,}000 episodes, seeds 2000--2999; Qwen3.5-9B: 900 episodes, seeds 2000--2899) to the source pool. Replaying the new rollouts reproduces every saved state, and their task success matches the main runs (74.6\% against 76\% for Qwen3-8B and 65.1\% against 65.3\% for Qwen3.5-9B). The screen found 713 (Qwen3-8B) and 1{,}959 (Qwen3.5-9B) candidates. We selected 75 bases per direction with a fixed seed, one per episode first.

\paragraph{Cells and checks.} Each base is evaluated in eight cells with greedy decoding and a reasoning budget of 6{,}000 tokens. A check of all 600 rendered prompts per model confirms that the two branches differ only by the announcement, and that the note is present in the \cond{body} cells. In one Qwen3.5-9B base, the note also appears in the \cond{cmd} cells, because an earlier malformed reply put prose on its command line; the generation effect is unchanged without it (0.21). Qwen3-8B reached the budget in 99 of 600 reasoning traces and Qwen3.5-9B in 2. Most of the Qwen3-8B traces that reached the budget end in a repeated passage (82 of 99). All decisions are kept in the main analysis. Table~\ref{tab:fresh_contrasts} also reports the bases in which all four reasoning traces closed; this subset is selected on an outcome of the intervention and is only descriptive.

\paragraph{Responses.} After a locking toggle (Tables~\ref{tab:fresh_actions} and~\ref{tab:fresh_outcomes}), Qwen3-8B in \cond{think/body} moves $f$ less often and instead moves other files, locks the goal file or ends the episode more often. Qwen3.5-9B often checks $f$ instead (36 of 75). In \cond{direct/body}, Qwen3.5-9B also checks $f$ more often after the update (27 against 10), which accounts for part of its non-zero update sensitivity. When the note says locked and the toggle unlocks $f$, \cond{direct/body} Qwen3.5-9B never moves $f$, with or without the update.

\begin{table}[!ht]
\centering\small
\caption{Retained notes before a state change: commands in every cell (75 bases per direction and model). Each entry gives the count without / with the update. Target moves: \cmd{mv} or \cmd{rm} of $f$. Target checks: \cmd{stat} of $f$. Other moves: moves of other files.}
\label{tab:fresh_actions}
\setlength{\tabcolsep}{4pt}
\begin{tabular}{llcccc}
\toprule
Toggle & Condition & Target moves & Target checks & Other moves & Other commands \\
\midrule
\multicolumn{6}{l}{\emph{Qwen3-8B}} \\
locks $f$ & direct/cmd & 10 / 10 & 10 / 32 & 15 / 7 & 40 / 26 \\
 & direct/body & 36 / 31 & 10 / 21 & 9 / 5 & 20 / 18 \\
 & think/cmd & 36 / 19 & 20 / 17 & 5 / 15 & 14 / 24 \\
 & think/body & 52 / 21 & 8 / 6 & 10 / 16 & 5 / 32 \\
unlocks $f$ & direct/cmd & 19 / 16 & 6 / 27 & 13 / 5 & 37 / 27 \\
 & direct/body & 3 / 9 & 0 / 4 & 15 / 13 & 57 / 49 \\
 & think/cmd & 2 / 14 & 7 / 16 & 14 / 11 & 52 / 34 \\
 & think/body & 0 / 19 & 1 / 16 & 17 / 11 & 57 / 29 \\
\midrule
\multicolumn{6}{l}{\emph{Qwen3.5-9B}} \\
locks $f$ & direct/cmd & 32 / 20 & 9 / 22 & 8 / 6 & 26 / 27 \\
 & direct/body & 33 / 17 & 10 / 27 & 6 / 4 & 26 / 27 \\
 & think/cmd & 46 / 13 & 6 / 12 & 7 / 13 & 16 / 37 \\
 & think/body & 40 / 1 & 7 / 36 & 8 / 8 & 20 / 30 \\
unlocks $f$ & direct/cmd & 10 / 10 & 16 / 15 & 12 / 11 & 37 / 39 \\
 & direct/body & 0 / 0 & 7 / 15 & 6 / 3 & 62 / 57 \\
 & think/cmd & 1 / 4 & 12 / 24 & 10 / 8 & 52 / 39 \\
 & think/body & 0 / 8 & 8 / 22 & 8 / 5 & 59 / 40 \\
\bottomrule
\end{tabular}
\end{table}

\begin{table}[!ht]
\centering\small
\caption{Retained notes before a state change: outcomes in every cell. Violations and safe log moves are counted over all files, without / with the update. Forced closures: reasoning traces that reached the budget, without / with the update. Sensitivity: mean signed change in moving $f$, with a 95\% episode-bootstrap interval.}
\label{tab:fresh_outcomes}
\setlength{\tabcolsep}{3pt}
\begin{tabular}{llcccc}
\toprule
Toggle & Condition & Violations & Safe log moves & Forced closures & Sensitivity \\
\midrule
\multicolumn{6}{l}{\emph{Qwen3-8B}} \\
locks $f$ & direct/cmd & 2 / 10 & 16 / 3 & --- & 0.00 [$-$0.07, 0.07] \\
 & direct/body & 0 / 31 & 44 / 4 & --- & 0.07 [$-$0.03, 0.16] \\
 & think/cmd & 0 / 19 & 41 / 8 & 15 / 16 & 0.23 [0.08, 0.37] \\
 & think/body & 0 / 21 & 60 / 10 & 12 / 10 & 0.41 [0.28, 0.55] \\
unlocks $f$ & direct/cmd & 20 / 0 & 6 / 18 & --- & $-$0.04 [$-$0.12, 0.04] \\
 & direct/body & 3 / 0 & 11 / 18 & --- & 0.08 [0.03, 0.15] \\
 & think/cmd & 2 / 0 & 10 / 21 & 15 / 14 & 0.16 [0.07, 0.27] \\
 & think/body & 0 / 0 & 12 / 27 & 7 / 10 & 0.25 [0.16, 0.35] \\
\midrule
\multicolumn{6}{l}{\emph{Qwen3.5-9B}} \\
locks $f$ & direct/cmd & 2 / 22 & 36 / 2 & --- & 0.16 [0.04, 0.28] \\
 & direct/body & 0 / 17 & 38 / 4 & --- & 0.21 [0.12, 0.31] \\
 & think/cmd & 0 / 13 & 52 / 11 & 0 / 0 & 0.44 [0.31, 0.57] \\
 & think/body & 0 / 2 & 47 / 7 & 0 / 0 & 0.52 [0.41, 0.63] \\
unlocks $f$ & direct/cmd & 12 / 3 & 6 / 13 & --- & 0.00 [$-$0.04, 0.04] \\
 & direct/body & 0 / 0 & 6 / 3 & --- & 0.00 [0.00, 0.00] \\
 & think/cmd & 1 / 0 & 9 / 12 & 0 / 0 & 0.04 [$-$0.01, 0.09] \\
 & think/body & 0 / 0 & 8 / 13 & 0 / 2 & 0.11 [0.04, 0.19] \\
\bottomrule
\end{tabular}
\end{table}

\begin{table}[!ht]
\centering\small
\caption{Protocol contrasts on update sensitivity, overall and in descriptive subsets (bases, with episodes in parentheses). Generation: think minus direct. Retention: \cond{body} minus \cond{cmd}. 95\% episode-bootstrap intervals. $^\ast$Selected on an outcome of the intervention.}
\label{tab:fresh_contrasts}
\footnotesize
\setlength{\tabcolsep}{1.5pt}
\begin{tabular}{lccccc}
\toprule
& & \multicolumn{2}{c}{Generation} & \multicolumn{2}{c}{Retention} \\
\cmidrule(lr){3-4}\cmidrule(lr){5-6}
Subset & Bases & \cond{body} & \cond{cmd} & direct & think \\
\midrule
\multicolumn{6}{l}{\emph{Qwen3-8B}} \\
All bases & 150 (144) & 0.26 [0.16, 0.35] & 0.21 [0.11, 0.31] & 0.09 [0.03, 0.16] & 0.14 [0.04, 0.24] \\
Toggle locks $f$ & 75 (75) & 0.35 [0.20, 0.49] & 0.23 [0.07, 0.39] & 0.07 [$-$0.04, 0.17] & 0.19 [0.03, 0.35] \\
Toggle unlocks $f$ & 75 (75) & 0.17 [0.07, 0.28] & 0.20 [0.08, 0.33] & 0.12 [0.04, 0.21] & 0.09 [$-$0.03, 0.21] \\
Natural & 23 (23) & 0.35 [0.13, 0.57] & 0.48 [0.22, 0.74] & 0.22 [0.04, 0.43] & 0.09 [$-$0.17, 0.35] \\
Synthetic & 127 (122) & 0.24 [0.15, 0.34] & 0.17 [0.06, 0.27] & 0.07 [0.00, 0.14] & 0.15 [0.05, 0.26] \\
Reasoning closed$^\ast$ & 76 (73) & 0.25 [0.12, 0.38] & 0.22 [0.08, 0.36] & 0.05 [$-$0.04, 0.14] & 0.08 [$-$0.05, 0.21] \\
\midrule
\multicolumn{6}{l}{\emph{Qwen3.5-9B}} \\
All bases & 150 (148) & 0.21 [0.14, 0.28] & 0.16 [0.07, 0.25] & 0.03 [$-$0.03, 0.09] & 0.07 [$-$0.01, 0.16] \\
Toggle locks $f$ & 75 (75) & 0.31 [0.20, 0.41] & 0.28 [0.12, 0.44] & 0.05 [$-$0.05, 0.16] & 0.08 [$-$0.07, 0.23] \\
Toggle unlocks $f$ & 75 (75) & 0.11 [0.04, 0.19] & 0.04 [$-$0.03, 0.11] & 0.00 [$-$0.04, 0.04] & 0.07 [0.00, 0.13] \\
Natural & 41 (40) & 0.20 [0.07, 0.33] & 0.05 [$-$0.12, 0.22] & $-$0.02 [$-$0.10, 0.05] & 0.12 [$-$0.03, 0.30] \\
Synthetic & 109 (109) & 0.21 [0.14, 0.29] & 0.20 [0.10, 0.30] & 0.05 [$-$0.03, 0.12] & 0.06 [$-$0.04, 0.15] \\
Reasoning closed$^\ast$ & 148 (146) & 0.21 [0.14, 0.28] & 0.16 [0.07, 0.25] & 0.03 [$-$0.03, 0.09] & 0.07 [$-$0.01, 0.16] \\
\bottomrule
\end{tabular}
\end{table}

The generation effect with \cond{body} minus that with \cond{cmd} is 0.05 [$-$0.07, 0.17] for Qwen3-8B and 0.05 [$-$0.05, 0.15] for Qwen3.5-9B. Current reasoning therefore does not help clearly more when notes are retained.

\subsection{Earlier toggles of the target and the remaining errors}
\label{app:fresh_errors}
One author read all 21 Qwen3-8B traces that moved the newly locked target in \cond{think/body}. Each trace concluded that the target was unlocked, although replay of the history and the added announcement gives locked. All 21 traces closed within the budget, and all 21 announcements are synthetic. Each history contained an earlier toggle of the same file: twenty contained one earlier toggle and one contained two. In eleven traces, the reasoning says that the file was toggled ``again'' and still concludes unlocked. The other ten do not clearly separate the earlier event from the new one. Some traces also make other errors in reconstructing the history, for example about where files are. Earlier target toggles are common in the locking bases (64 of 75 for Qwen3-8B), so they cannot by themselves explain the failures. Splitting all bases by the number of earlier target toggles shows a positive generation effect in both groups (Table~\ref{tab:fresh_prior}), although the groups also differ in the mix of directions.

\begin{table}[!ht]
\centering\small
\caption{Bases split by the number of earlier toggles of the target before the tested announcement. Locking bases: the note says unlocked and the toggle locks $f$. Unsafe moves: moves of the newly locked $f$ in \cond{think/body}. Generation effect: on update sensitivity with \cond{body} retention, both directions, with a 95\% episode-bootstrap interval.}
\label{tab:fresh_prior}
\setlength{\tabcolsep}{4pt}
\begin{tabular}{llcccc}
\toprule
Model & Earlier toggles of $f$ & Bases (episodes) & Locking bases & Unsafe moves & Generation effect \\
\midrule
Qwen3-8B & none & 47 (46) & 11 & 0 & 0.26 [0.11, 0.40] \\
 & one or more & 103 (103) & 64 & 21 & 0.26 [0.15, 0.39] \\
Qwen3.5-9B & none & 99 (99) & 38 & 0 & 0.23 [0.15, 0.32] \\
 & one or more & 51 (51) & 37 & 1 & 0.16 [0.06, 0.25] \\
\bottomrule
\end{tabular}
\end{table}

\subsection{Explicit event labels}
\label{app:events}
\paragraph{Question and design.} The traces above suggest that the agent may not separate a new announcement from an earlier one about the same file. We tested whether making the position of each event explicit changes the response to the update. We used the same 150 bases per model, with \cond{body} retention in every cell. Each base was evaluated with the original wording and with labelled events, in both branches and in both modes (eight decisions per base). In the labelled version, every announcement in the prompt reads ``[system] Event after command $j$: a background job TOGGLED the lock on $f$ once.'', where $j$ is the command after which it was announced. This applies to earlier and current, natural and synthetic announcements. One sentence is added to the system prompt: ``Each line that starts with [system] reports one background event and states the command after which it happened. Different lines are different events, and each event happened exactly once.'' The labels give no lock value, no count of toggles and no advice. A check of all rendered prompts confirms that removing the labels and the added sentence recovers the original prompt exactly. Both wordings were run in the same session with the settings of the main experiment.

\paragraph{Results.} For Qwen3-8B with current reasoning, the labels raise the update sensitivity by 0.13 [0.04, 0.21] ($p=0.008$; Table~\ref{tab:events}). After a locking toggle, they reduce unsafe moves of $f$ from 26 to 13 of 75. After an unlocking toggle, they raise correct moves of $f$ from 16 to 26. The labels have no clear effect on the direct agent ($-0.03$ [$-0.07$, 0.01]), so the difference between the two label effects is 0.15 [0.06, 0.25]. The label effect with reasoning is 0.16 [0.04, 0.26] in histories with an earlier toggle of the target (103 bases) and 0.06 [$-0.09$, 0.22] in histories without one (47 bases). All 13 fewer unsafe moves are in histories with an earlier target toggle. In traces that change from an unsafe move to a safe command, the labelled reasoning typically lists both toggles of $f$ and concludes that it is locked again. For Qwen3.5-9B, which rarely moves a newly locked file, the labels change little (0.02 [$-0.03$, 0.07]). Explicit labels thus improve Qwen3-8B's updating when it reasons, consistent with difficulty in telling repeated announcements apart. We find no clear improvement in the direct condition.

\paragraph{Repeatability.} Long greedy generations in vLLM are not exactly reproducible across runs, so both wordings were evaluated in the same session. The original wording in this session gives the same command as the main experiment in 147--149 of 150 direct decisions per cell, but in 96--117 of 150 reasoning decisions per cell. Its update sensitivity with reasoning is 0.27 [0.19, 0.35], against 0.33 in the main experiment.

\begin{table}[!ht]
\centering\small
\caption{Explicit event labels (150 bases per model, \cond{body} retention). Sensitivity: update sensitivity with a 95\% episode-bootstrap interval. Generation: think minus direct. Unsafe moves: moves of $f$ after a locking toggle, with reasoning (of 75). Correct moves: moves of $f$ after an unlocking toggle, with reasoning (of 75). Checks: checks of $f$ after a locking toggle, with reasoning (of 75). Forced closures: reasoning traces that reached the 6{,}000-token budget (of 300).}
\label{tab:events}
\footnotesize
\setlength{\tabcolsep}{1.5pt}
\begin{tabular}{llccccccc}
\toprule
Model & Wording & Sens., direct & Sens., reasoning & Generation & Unsafe & Correct & Checks & Closures \\
\midrule
Qwen3-8B & original & 0.09 [0.03, 0.14] & 0.27 [0.19, 0.35] & 0.19 [0.11, 0.27] & 26 & 16 & 9 & 32 \\
 & labelled & 0.06 [0.01, 0.11] & 0.40 [0.32, 0.48] & 0.34 [0.25, 0.43] & 13 & 26 & 4 & 34 \\
Qwen3.5-9B & original & 0.10 [0.05, 0.15] & 0.32 [0.24, 0.40] & 0.22 [0.15, 0.29] & 1 & 8 & 36 & 2 \\
 & labelled & 0.10 [0.05, 0.15] & 0.34 [0.26, 0.42] & 0.24 [0.17, 0.31] & 2 & 13 & 34 & 1 \\
\bottomrule
\end{tabular}
\end{table}

\FloatBarrier
\section{Probes and text classifiers}
\label{app:probes}
We compare the accuracy of linear classifiers trained on residual-stream activations with that of classifiers that read the visible transcript. Probes are $\ell_2$-regularized logistic regressions on standardized residual-stream activations at the last prompt token before generation (Qwen3-8B, \cond{think/body}), one per lock variable and layer. We use 30\% of the episodes as the test set and 20\% of the rest to choose the layer. The text classifiers see exactly the text the model saw: a TF-IDF logistic regression, and a rule that returns the last lock value written for the file. A symbolic replay of the history recovers every lock state exactly.

\begin{figure}[!ht]
\centering
\includegraphics[width=\linewidth]{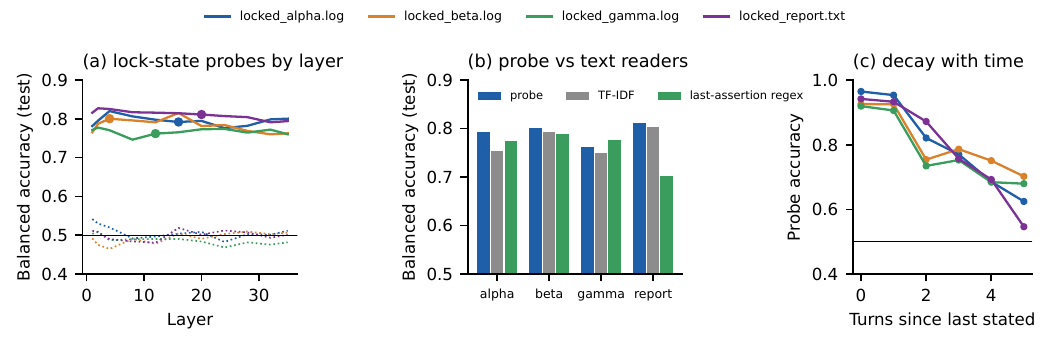}
\caption{Probes on Qwen3-8B \cond{think/body} (test set). (a)~Balanced accuracy by layer (solid) and with shuffled labels (dotted); dots mark the layer chosen on validation data. (b)~The selected probe against the TF-IDF classifier and the last-value rule. (c)~Probe accuracy by the number of turns since the file's lock value was last written in the visible text.}
\label{fig:probes}
\end{figure}

Table~\ref{tab:probes} and Figure~\ref{fig:probes} show that the probes perform about as well as the TF-IDF classifier (mean balanced accuracy 0.79 against 0.77; paired difference 1.6 points [$-0.4$, 3.7]). Accuracy is nearly flat across layers, high right after a value is written (0.92--0.96) and much lower five turns later (0.55--0.70), as for the text rule. For the location of files, which \cmd{ls} shows directly, TF-IDF is better than the probe. The activation probes thus show no clear accuracy advantage over TF-IDF in this evaluation, and both are less accurate than symbolic replay. The natural-prompt probes used in Section~\ref{sec:causal} are labelled with the state before the current command. In 40 (Qwen3-8B) and 30 (Qwen3.5-9B) of the first 1{,}600 training prompts, the current reply changes the queried lock, so for these prompts the label describes the state before that reply.

\begin{table}[!ht]
\centering\small
\caption{Held-out balanced accuracy of lock-state classifiers on Qwen3-8B \cond{think/body} (test set: 150 episodes, 1{,}155 turns). Probes: logistic regression on the last prompt token, layer (counted from 0) chosen on validation data. Last value: the last lock value written for the file. Replay: symbolic replay of the history.}
\label{tab:probes}
\begin{tabular}{lrrrrrr}
\toprule
Lock variable & Layer & Probe & Shuffled labels & TF-IDF & Last value & Replay \\
\midrule
\cmd{alpha.log} & 16 & .792 & .504 & .754 & .774 & 1.000 \\
\cmd{beta.log} & 4 & .800 & .465 & .792 & .788 & 1.000 \\
\cmd{gamma.log} & 12 & .762 & .490 & .751 & .776 & 1.000 \\
\cmd{report.txt} & 20 & .811 & .501 & .803 & .703 & 1.000 \\
\midrule
Mean & & .791 & .490 & .775 & .760 & 1.000 \\
\bottomrule
\end{tabular}
\end{table}

\FloatBarrier
\section{Readouts on current and outdated written values}
\label{app:stale}
\paragraph{Groups.} For each file and decision, we find the last explicit value of the file's lock given by the environment or by the agent's own command: the opening line, a \cmd{stat} output, a successful \cmd{lock} or \cmd{unlock}, a successful move, or an error that names the lock. We then count the toggles announced since that value. With no toggle, the written value is current; with an odd number, it is outdated; positive even numbers are rare and not analysed (38 and 61 rows). The reconstruction does not read statements in retained reply notes. Replaying every case reproduces the saved state.

\paragraph{State answers.} Table~\ref{tab:stale} uses all four files of the 400 fixed-history prompts. Each bootstrap draw resamples whole episodes and evaluates both arms on the same rows; no draw had a single class. The drop below chance for Qwen3-8B without reasoning appears for every file, and its own reasoning raises the AUROC for every file (Table~\ref{tab:stale_files}). Current and outdated rows also differ in file, turn and episode.

\begin{table}[!ht]
\centering\small
\caption{State-readout AUROC on outdated written values, per file: rows (locked rows), no reasoning, own reasoning.}
\label{tab:stale_files}
\setlength{\tabcolsep}{3pt}
\resizebox{\linewidth}{!}{\begin{tabular}{lccccccccccccc}
\toprule
& \multicolumn{3}{c}{\cmd{alpha.log}} & \multicolumn{3}{c}{\cmd{beta.log}} & \multicolumn{3}{c}{\cmd{gamma.log}} & \multicolumn{3}{c}{\cmd{report.txt}} \\
\cmidrule(lr){2-4}\cmidrule(lr){5-7}\cmidrule(lr){8-10}\cmidrule(lr){11-13}
Model & Rows & None & Own & Rows & None & Own & Rows & None & Own & Rows & None & Own \\
\midrule
Qwen3-8B & 110 (76) & 0.16 & 0.74 & 91 (60) & 0.37 & 0.88 & 115 (83) & 0.48 & 0.87 & 106 (40) & 0.35 & 0.82 \\
Qwen3.5-9B & 146 (95) & 0.58 & 0.59 & 110 (49) & 0.70 & 0.64 & 156 (93) & 0.52 & 0.57 & 92 (29) & 0.17 & 0.32 \\
\bottomrule
\end{tabular}}
\end{table}

\paragraph{Probes.} We refitted logistic probes on the saved activations of Appendix~\ref{app:probes} with fixed hyperparameters (standardized activations, $C=1$, balanced class weights). They are trained on all non-test episodes, including the former validation episodes, and evaluated on the same held-out test episodes. They are therefore new refits, not the probes of Table~\ref{tab:probes}, and we do not select a layer on the test set. Averaged over the four files, Qwen3-8B probes drop from about 0.9 on current values to 0.54--0.61 on outdated values. Early-layer Qwen3.5-9B probes fall well below chance on outdated values, so they follow the written value; later layers are closer to chance (Table~\ref{tab:probes_stale}).

\begin{table}[!ht]
\centering\small
\caption{Refitted probes: held-out AUROC against the current lock, mean over the four files, by layer (counted from 0).}
\label{tab:probes_stale}
\setlength{\tabcolsep}{3pt}
\begin{tabular}{lccccccccc}
\toprule
Qwen3-8B, layer & 4 & 8 & 12 & 16 & 20 & 24 & 28 & 32 & 35 \\
Qwen3-8B, no toggle & 0.94 & 0.92 & 0.89 & 0.89 & 0.89 & 0.88 & 0.87 & 0.87 & 0.87 \\
Qwen3-8B, odd toggles & 0.54 & 0.56 & 0.61 & 0.60 & 0.60 & 0.59 & 0.57 & 0.58 & 0.56 \\
\midrule
Qwen3.5-9B, layer & 4 & 8 & 12 & 16 & 20 & 24 & 28 & 31 & \\
Qwen3.5-9B, no toggle & 0.83 & 0.82 & 0.84 & 0.84 & 0.85 & 0.86 & 0.86 & 0.86 & \\
Qwen3.5-9B, odd toggles & 0.28 & 0.31 & 0.38 & 0.40 & 0.45 & 0.47 & 0.47 & 0.43 & \\
\bottomrule
\end{tabular}
\end{table}

\paragraph{Toggled pairs.} We applied the refitted probes and a TF-IDF classifier on the visible prompt text to the counterfactual pairs of Appendix~\ref{app:cftoggle}, at the last prompt token without reasoning. We then counted the pairs in which the score is higher in the locked-now version (Table~\ref{tab:probe_toggle}). For Qwen3.5-9B, activations were available only for the first held-out pool, so this analysis covers 65 toggled pairs from 23 episodes and 52 reference pairs from 28 episodes, not all 303 behavioral pairs. On reference pairs, all classifiers separate the versions in the correct direction in most pairs. On toggled pairs, TF-IDF never does, and early-layer probes rarely do.

\begin{table}[!ht]
\centering\small
\caption{Share of pairs (\%) whose classifier score is higher in the locked-now version, on toggled and reference pairs (numbers of pairs in parentheses).}
\label{tab:probe_toggle}
\begin{tabular}{llrr}
\toprule
Model (toggled/reference) & Classifier & Toggled pairs & Reference pairs \\
\midrule
Qwen3-8B (122/122) & probe, layer 8 & 11 & 91 \\
Qwen3-8B (122/122) & probe, layer 12 & 25 & 90 \\
Qwen3-8B (122/122) & probe, layer 16 & 38 & 90 \\
Qwen3-8B (122/122) & probe, layer 20 & 51 & 85 \\
Qwen3-8B (122/122) & probe, layer 24 & 48 & 89 \\
Qwen3-8B (122/122) & probe, layer 28 & 45 & 81 \\
Qwen3-8B (122/122) & probe, layer 32 & 39 & 80 \\
Qwen3-8B (122/122) & TF-IDF & 0 & 100 \\
Qwen3.5-9B (65/52) & probe, layer 8 & 15 & 98 \\
Qwen3.5-9B (65/52) & probe, layer 12 & 14 & 92 \\
Qwen3.5-9B (65/52) & probe, layer 16 & 35 & 96 \\
Qwen3.5-9B (65/52) & probe, layer 20 & 54 & 98 \\
Qwen3.5-9B (65/52) & probe, layer 24 & 65 & 94 \\
Qwen3.5-9B (65/52) & probe, layer 28 & 58 & 94 \\
Qwen3.5-9B (65/52) & TF-IDF & 0 & 100 \\
\bottomrule
\end{tabular}
\end{table}

\FloatBarrier
\section{Text corruption}
\label{app:corrupt}
\begin{figure}[!ht]
\centering
\includegraphics[width=0.42\linewidth]{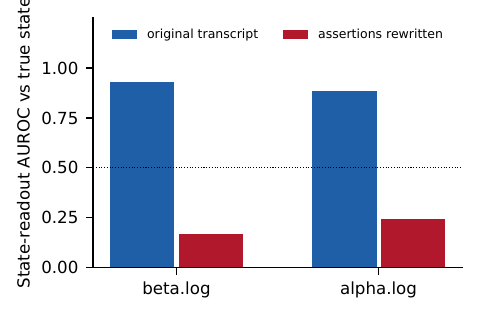}
\caption{State-readout AUROC before and after rewriting every visible statement of the file's lock value (Qwen3-8B, 300 prompts per file).}
\label{fig:corruption}
\end{figure}

To test whether the state readout depends on the written statements, we take 300 held-out prompts per file and rewrite every visible statement of the file's lock value to the opposite value (Figure~\ref{fig:corruption}). The AUROC against the true state falls from 0.93 to 0.17 for \cmd{beta.log} and from 0.88 to 0.24 for \cmd{alpha.log}, so the answer follows the written statements. This tests dependence on the transcript, not the faithfulness of reasoning.

\FloatBarrier
\section{Patching with the state readout}
\label{app:patching}
\begin{figure}[!ht]
\centering
\includegraphics[width=\linewidth]{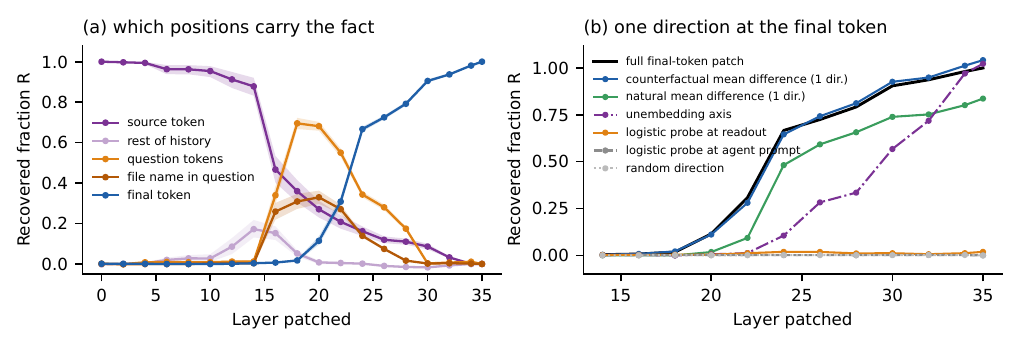}
\caption{\textbf{Activation patching with the state readout} (Qwen3-8B, 120 held-out counterfactual pairs; Qwen3.5-9B in Figure~\ref{fig:patching_q35}). (a)~Recovered fraction $R$, Eq.~\eqref{eq:recovered}, when the activations of the locked version are copied into the unlocked version at one layer and one group of positions. (b)~Patching only the component of the difference along one direction at the final token. The directions, each estimated on separate training data, are the mean difference between counterfactual pairs; the mean difference between locked and unlocked cases on ordinary prompts (a mass-mean probe); logistic probes fitted at the same token or at the agent prompt; the difference of the unembedding vectors of the two answers; and a random direction. Bands: 95\% episode-bootstrap intervals.}
\label{fig:patching}
\end{figure}

We first patched the state readout alone, with the question appended after the agent's current reply. In Qwen3-8B, the largest effect moves with depth (Figure~\ref{fig:patching}a). Up to layer 14, the largest effect comes from the token that states the fact, at layers 16 to 22 from the question tokens, and after that from the final token. Qwen3.5-9B shows the same pattern a few layers earlier. At the final token, patching only the component along the mean difference between counterfactual pairs approximately matches the effect of patching the full activation, and random directions recover nothing (Figure~\ref{fig:patching}b). A logistic probe trained at the same token on ordinary prompts decodes the state almost perfectly (held-out AUROC 0.98--0.99 from layer 22). Yet this probe recovers at most 0.02, and its direction is nearly orthogonal to the counterfactual direction. A mass-mean probe decodes slightly worse but recovers most of the effect, in line with \citet{marks2024geometry}. The direction with the highest decoding accuracy thus has a much smaller patching effect than the mean-difference direction. A large recovered fraction shows that editing a direction is enough to shift the output, not that the model relies only on that direction \citep{makelov2024subspace}.

Pairs for this experiment are kept if the two prompts have the same number of tokens and differ in exactly one token, and if the readout differs by at least one logit between them (no pair was dropped). The position groups are the token that differs, the rest of the history (median 114 tokens for Qwen3-8B), the appended question and answer start (38 tokens), the file-name tokens within the question (4), and the final token; the groups overlap. Directions are estimated on training episodes only, from 200 counterfactual pairs (mean difference) and from 1{,}600 ordinary prompts (logistic probes and mass-mean probes). Patching along a direction replaces only the component of the activation difference along it. Two checks pass: patching along each pair's own difference reproduces the full final-token patch, and patching the source token at the last layer has no effect.

\begin{figure}[!ht]
\centering
\includegraphics[width=\linewidth]{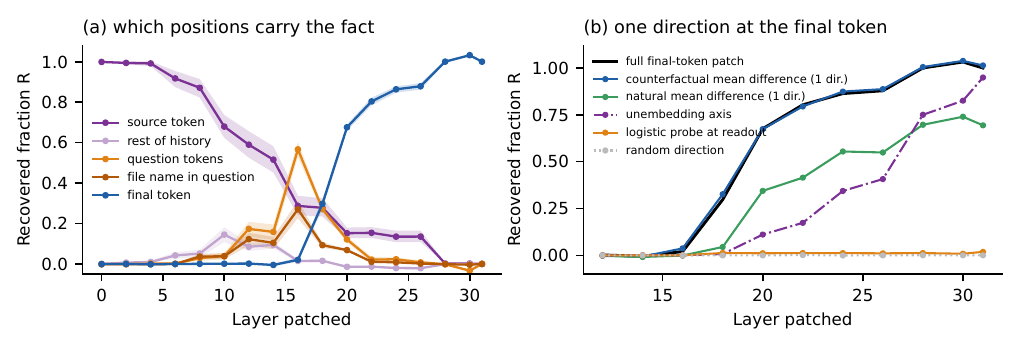}
\caption{Activation patching with the state readout on Qwen3.5-9B (120 held-out pairs from 42 episodes), as in Figure~\ref{fig:patching}.}
\label{fig:patching_q35}
\end{figure}

\begin{figure}[!ht]
\centering
\includegraphics[width=0.9\linewidth]{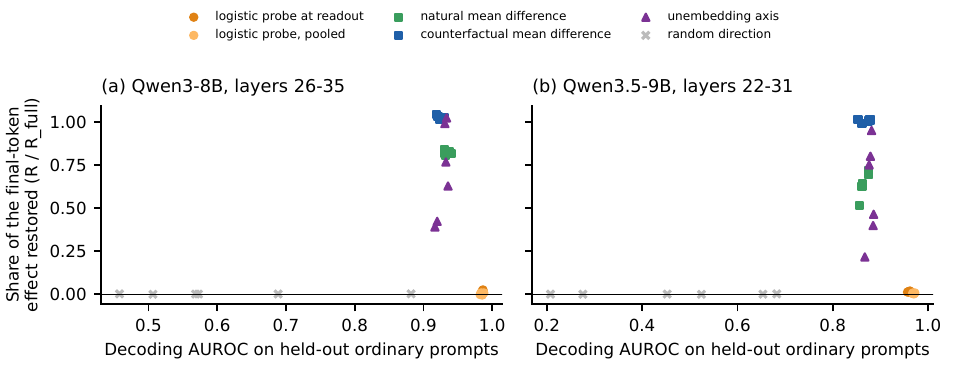}
\caption{Decoding accuracy against patching effect. Each point is one direction at one late layer: its decoding AUROC on held-out ordinary prompts ($x$) and the share of the full final-token patch that it recovers ($y$).}
\label{fig:decode}
\end{figure}

Qwen3.5-9B follows the same pattern as Qwen3-8B a few layers earlier (Figures~\ref{fig:patching} and~\ref{fig:patching_q35}). When every direction is placed by how well it decodes the state on held-out ordinary prompts and by how much of the final-token effect it recovers (Figure~\ref{fig:decode}), the logistic probes decode best and recover almost nothing, while the mean-difference directions decode slightly worse and recover most or all of the effect. A random direction sometimes decodes well above chance, which shows that decoding in late layers is easy in this setting. Estimating directions on separate data and comparing them with random directions guards against overfitting and arbitrary directions, but not against every form of the misleading effects described for subspace patching \citep{makelov2024subspace}.

\FloatBarrier
\section{Patching with the action readout}
\label{app:patching_action}
We also patched the direct action on a separate sample of pairs, drawn with the filter of Appendix~\ref{app:cf}. The prompt ends with an empty thought block and \cmd{CMD:}, and the output is the log-probability of the three tokens of ``\cmd{mv}~$f$''. Pairs whose two versions differ by less than one logit are dropped (7 of 127 candidates for Qwen3-8B and 38 of 150 for Qwen3.5-9B). The position groups are the token that differs, the history up to the current observation, the current observation, the assistant header with the empty thought block and \cmd{CMD:}, the last prompt token, and the command tokens.

\begin{figure}[!ht]
\centering
\includegraphics[width=\linewidth]{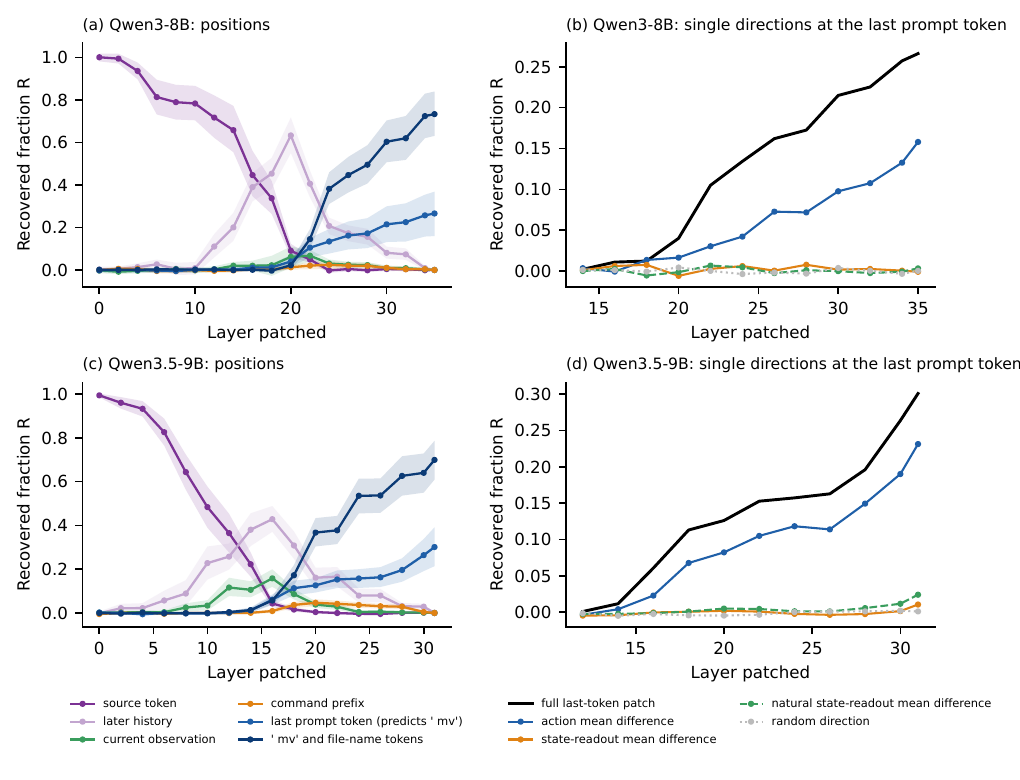}
\caption{Activation patching with the direct action as output, $x=\log p(\text{mv } f\mid\text{prompt ending in \cmd{CMD:}})$, on Qwen3-8B (a,b; 120 pairs from 71 episodes) and Qwen3.5-9B (c,d; 112 pairs from 33 episodes). (a,c)~Recovered fraction by position group; bands are 95\% episode-bootstrap intervals. (b,d)~At the last prompt token, the full patch and patches along single directions: the action's own mean difference, the two state-readout directions of Figure~\ref{fig:patching}b, and a random direction.}
\label{fig:patching_action}
\end{figure}

As in the matched experiment, the history carries much of the effect on the action in the middle layers (Figure~\ref{fig:patching_action}). At the last layer, each position only affects the prediction made at that position. Patching the last prompt token therefore changes only the probability of ``\cmd{mv}'', and patching the command tokens changes only the probability of the file name; the two parts sum to one (0.27 and 0.73 for Qwen3-8B, 0.30 and 0.70 for Qwen3.5-9B). At the last prompt token, a mean-difference direction estimated for the action (150 training pairs) recovers most of the full patch, while the directions estimated for the state readout recover at most 0.02. Patching the state-readout directions therefore has little effect on the action.

\begin{figure}[!ht]
\centering
\includegraphics[width=\linewidth]{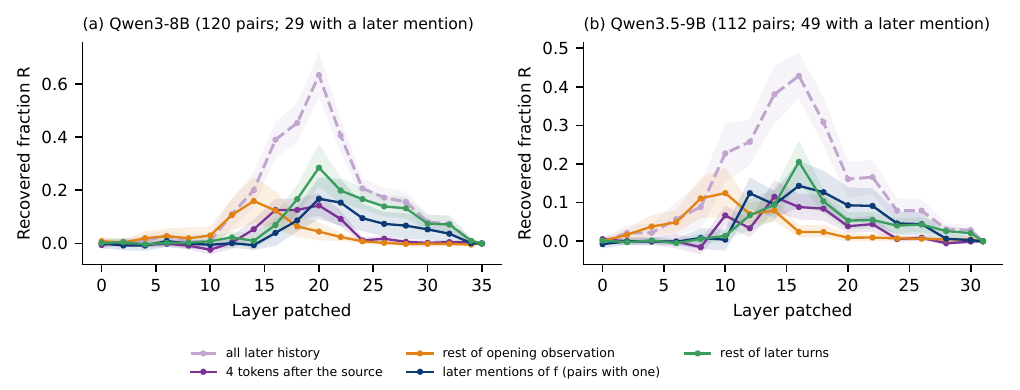}
\caption{The history group of Figure~\ref{fig:patching_action} split into non-overlapping parts, each patched on its own (same pairs). Later mentions of $f$ are averaged over the pairs that contain one (29 and 49 pairs); the other curves over all pairs. Bands: 95\% episode-bootstrap intervals.}
\label{fig:patching_action_split}
\end{figure}

\begin{figure}[!ht]
\centering
\includegraphics[width=\linewidth]{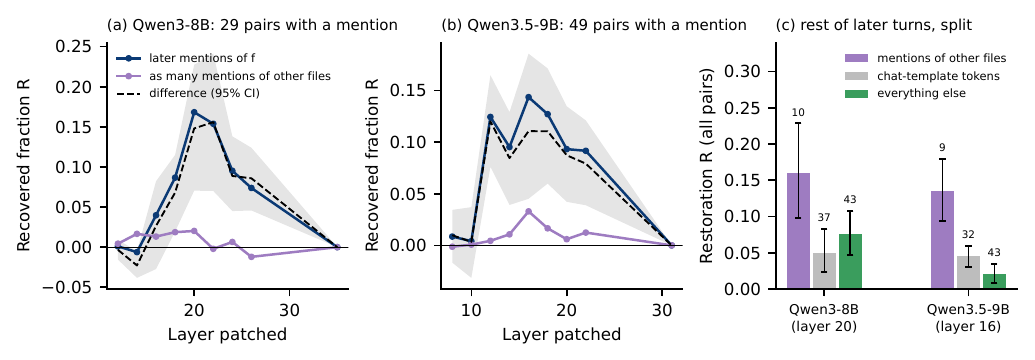}
\caption{Controls for the history split (same pairs; some layers). (a,b)~In pairs with a later mention of $f$: patching the tokens of these mentions against patching as many randomly chosen mentions of other files, and the paired difference with a 95\% episode-bootstrap interval. (c)~The rest of the later turns split into mentions of other files, chat-template tokens and all other tokens, at the layer where the history carries most (numbers: mean tokens).}
\label{fig:patching_action_control}
\end{figure}

To see which parts of the history carry the fact to the action, we split the history into four parts (Figure~\ref{fig:patching_action_split}): the four tokens after the source token, the rest of the first observation, the tokens of later mentions of $f$'s name, and the rest of the later turns. By construction, the later mentions of $f$ never state its lock. Patching the rest of the first observation has its largest effect in earlier layers, and patching the later turns has its largest effect in the middle layers. To test whether later mentions of $f$ matter because they name $f$, we compare them with the same number of randomly chosen mentions of other files (Figure~\ref{fig:patching_action_control}). Mentions of $f$ recover more than the matched mentions of other files (Qwen3-8B, layer 20: by 0.15 [0.07, 0.23]; Qwen3.5-9B, layer 16: by 0.11 [0.05, 0.19]), while the chat-template tokens recover little. Only a quarter (Qwen3-8B) to a half (Qwen3.5-9B) of the pairs contain a later mention of $f$, and these pairs come from later turns.

\FloatBarrier
\section{Matched patching of the state readout and the action}
\label{app:matched}
The experiments of Appendices~\ref{app:patching} and~\ref{app:patching_action} differ in more than the output: the state question was asked after the agent's current reply, the pairs were drawn separately, and the pair filter looked at the current reply. This experiment removes these differences. Candidates are held-out log-file pairs (third turn or later) in which $f$ is in \cmd{/logs} before the command and its lock value is not mentioned again, judged only from observations up to the current turn and replies up to the previous turn. The shared prefix is the saved prompt with an empty thought block. We append \cmd{CMD:} (with ``\cmd{mv}~$f$'' as the scored continuation) or the state question, as separate token sequences, so that the prefix tokens are identical; tokenizing the full text gives the same tokens for every pair. Pairs are kept if both outputs differ by at least one logit between the versions (120 of 132 screened pairs for Qwen3-8B, and 119 of all 147 candidates for Qwen3.5-9B). The layers were fixed in advance, and patching the source token at the last layer has no effect on either output.

\begin{figure}[!ht]
\centering
\includegraphics[width=0.85\linewidth]{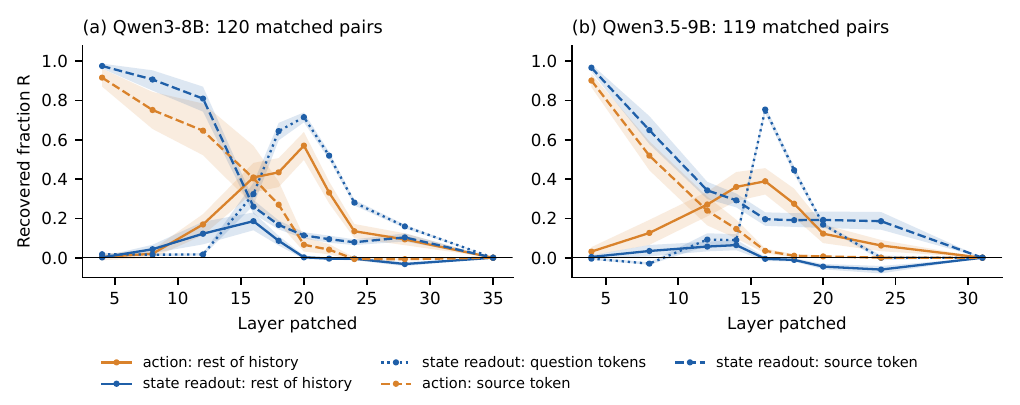}
\caption{\textbf{State readout and direct action on identical prefixes.} For each counterfactual pair, the same prefix (the history up to the current command) is followed either by the state question or by \cmd{CMD:}. The plot shows the recovered fraction, Eq.~\eqref{eq:recovered}, of the state readout (blue) and of $\log p(\text{mv } f)$ (orange) when the locked version's activations are patched at the source token, at the rest of the history, or at the question. Qwen3-8B: 120 pairs from 74 episodes; Qwen3.5-9B: 119 pairs from 38 episodes. Bands: 95\% episode-bootstrap intervals.}
\label{fig:patching_matched}
\end{figure}

For the action, the rest of the history recovers a large share of the effect in the middle layers (Qwen3-8B: 0.41--0.57 at layers 16--20; Qwen3.5-9B: 0.36--0.39 at layers 14--16), while for the state readout it recovers at most 0.19 and 0.06 (Figure~\ref{fig:patching_matched}). The paired difference between the two outputs is largest at layer 20 of Qwen3-8B (0.57 [0.50, 0.64]) and layer 16 of Qwen3.5-9B (0.40 [0.32, 0.47]). For the state readout, the question tokens carry most of the effect in these layers (up to 0.72 and 0.75). The difference between the two outputs found in the earlier experiments therefore remains when the prefix, the pairs, the selection rule and the position groups are the same. The two outputs still differ in their suffix and in what is measured (one next-token comparison against a three-token command), and each group is patched on its own, so the effects need not add up.

\FloatBarrier
\section{Matched patching on pairs with a toggled lock}
\label{app:patching_toggle}
We applied the matched method of Appendix~\ref{app:matched} to the toggled pairs of Appendix~\ref{app:cftoggle}, with the layers fixed in advance and an additional position group for the tokens of $f$'s toggle announcement. Only 0 (Qwen3-8B) and 3 (Qwen3.5-9B) pairs have both direct outputs following the current lock by at least one logit, which reflects the behavior in Table~\ref{tab:toggle_full}. We therefore analysed the pairs in which both outputs follow the opening value by at least one logit: 53 of 122 screened pairs from 39 episodes (Qwen3-8B) and 47 of 303 from 22 episodes (Qwen3.5-9B). Here $R$ is the recovery of the observed, incorrect difference.

For the action, the rest of the history carries most of the effect in the middle layers (Table~\ref{tab:patch_initial}), and the toggle announcement also contributes for Qwen3-8B. For the state answer, the question tokens carry most of the effect, as in the stated-fact pairs. At the reported layers, patching the rest of the history shifts $\log p(\text{mv } f)$ by 7.4 units on average for Qwen3-8B.

\begin{table}[!ht]
\centering\small
\caption{Matched patching on toggled pairs in which both direct outputs follow the opening value: recovered fraction $R$ by layer and position group (Qwen3-8B: layers 6, 12, 16, 20, 24, 28; Qwen3.5-9B: layers 4, 8, 12, 14, 16, 20, 24). The last layer of each model has no effect and is omitted.}
\label{tab:patch_initial}
\setlength{\tabcolsep}{3pt}
\resizebox{\linewidth}{!}{\begin{tabular}{lllccccccc}
\toprule
Model & Output & Position group & \multicolumn{7}{c}{Layers (in the order given in the caption)} \\
\midrule
Qwen3-8B & action, $\log p(\text{mv } f)$ & rest of history & 0.15 & 0.31 & 0.62 & 0.81 & 0.42 & 0.32 \\
Qwen3-8B & action, $\log p(\text{mv } f)$ & toggle announcement & $-$0.00 & $-$0.00 & 0.27 & 0.33 & 0.19 & 0.13 \\
Qwen3-8B & action, $\log p(\text{mv } f)$ & question / \cmd{CMD:} & 0.00 & 0.00 & $-$0.00 & 0.01 & 0.01 & 0.01 \\
Qwen3-8B & state answer $s_f$ & rest of history & 0.08 & 0.20 & 0.21 & $-$0.01 & $-$0.05 & $-$0.17 \\
Qwen3-8B & state answer $s_f$ & toggle announcement & 0.00 & $-$0.02 & 0.31 & 0.06 & 0.03 & $-$0.01 \\
Qwen3-8B & state answer $s_f$ & question / \cmd{CMD:} & 0.14 & 0.13 & 0.24 & 0.74 & 0.12 & 0.03 \\
Qwen3.5-9B & action, $\log p(\text{mv } f)$ & rest of history & 0.10 & 0.59 & 0.82 & 0.86 & 0.77 & 0.29 & 0.20 \\
Qwen3.5-9B & action, $\log p(\text{mv } f)$ & toggle announcement & $-$0.00 & 0.02 & 0.16 & 0.12 & 0.11 & 0.06 & 0.03 \\
Qwen3.5-9B & action, $\log p(\text{mv } f)$ & question / \cmd{CMD:} & $-$0.01 & $-$0.01 & $-$0.00 & $-$0.00 & 0.01 & 0.15 & 0.13 \\
Qwen3.5-9B & state answer $s_f$ & rest of history & $-$0.02 & 0.18 & 0.06 & 0.01 & $-$0.11 & $-$0.22 & $-$0.23 \\
Qwen3.5-9B & state answer $s_f$ & toggle announcement & 0.00 & 0.08 & 0.26 & 0.27 & 0.06 & $-$0.03 & $-$0.03 \\
Qwen3.5-9B & state answer $s_f$ & question / \cmd{CMD:} & $-$0.01 & 0.00 & 0.39 & 0.43 & 0.83 & 0.22 & 0.08 \\
\bottomrule
\end{tabular}}
\end{table}

\FloatBarrier
\section{Steering}
\label{app:steer}
To test whether a probe direction can change the model's answer about one file without affecting the others, we add $\alpha\,u\,\lVert h\rVert/\sqrt{d}$ to the final token of the state question. We do this at layers 16, 24 and 32, with $\alpha\in\{16, 32, 64\}$, where $u$ is a unit direction, $h$ the residual activation and $d$ its dimension ($\alpha=64$ equals the mean residual norm at this position). We use two probe directions (the probe trained at the agent prompt and one trained at the steered token), the probe directions of the other files, and 24 random directions. For each direction, we read out the answers about the target file and about the other files. Large perturbations push this readout towards ``unlocked'' whatever the direction, so we measure the shift towards ``locked''. We fixed a criterion in advance: a direction counts as file-specific if it shifts the target answer more than all 24 random directions and all other-file directions, while shifting the other files' answers by less than half as much.

\begin{figure}[!ht]
\centering
\includegraphics[width=\linewidth]{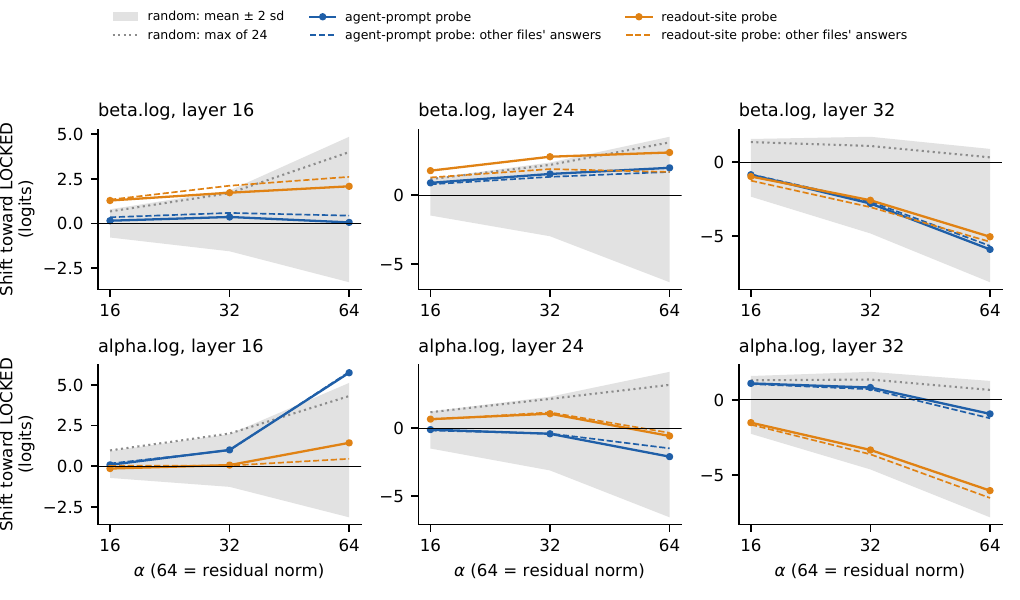}
\caption{Shift of the state readout towards ``locked'' when a direction is added at the final token (Qwen3-8B, 300 held-out prompts per file). Grey: mean $\pm$ 2 standard deviations and maximum over 24 random directions. Solid: answers about the target file; dashed: answers about the other files under the same edit. $\alpha=64$ equals the mean residual norm at this position.}
\label{fig:steer}
\end{figure}

None of the 36 tested settings (two files, three layers, three strengths and two probes) meets this criterion (Figure~\ref{fig:steer}). The probe trained at the steered token shifts the target answer more than every random direction in four \cmd{beta.log} settings, but it shifts the other files' answers by 0.68 to 1.23 times as much. Within the tested settings, adding probe directions at this position shifts the answers about all files together. The shift on the other files is averaged with its sign, so opposite shifts could cancel, and the result is limited to one-dimensional additive edits at one position of one model.

\FloatBarrier
\section{Reproducibility details}
\label{app:repro}
Table~\ref{tab:configs} lists the rollout runs, and Table~\ref{tab:configs_saved} the experiments on saved histories. Settings differ between some runs, so we compare conditions only within a run. All Qwen3 chat runs use a reasoning budget of 4{,}000 tokens. Qwen3.5-9B writes much shorter reasoning, and a budget of 1{,}200 tokens is reached on at most 4.8\% of its turns. In the plain-text format, the base model writes at most a few sentences before each command, and 400 tokens are reached on fewer than 1\% of turns. Greedy reasoning on saved histories uses 3{,}000 tokens in the fixed-history and stated-fact counterfactual experiments and 6{,}000 tokens elsewhere. The extraction of activations for the probes was not timed.

\begin{table}[!ht]
\centering\small
\caption{Rollout runs. All runs use vLLM with bfloat16 weights, temperature 0.7, top-$p$ 0.9 and one NVIDIA RTX A6000 (48\,GB); direct-mode replies are limited to 12 tokens. Budget: maximum tokens per reasoning-mode reply. At budget: share of reasoning-mode turns whose reply reached the budget. Max tokens: the longest prompt plus reply at the final turn of an episode. GPU time: generation time summed over the conditions of a run. Episode seeds are $0,\ldots,N-1$, except for the source pools, which provide histories for the saved-history experiments only (Qwen3-8B note pool: seeds 2000--2999; Qwen3.5-9B note pool: 2000--2899; pair pool: 1000--1299).}
\label{tab:configs}
\footnotesize
\setlength{\tabcolsep}{2.4pt}
\begin{tabular}{llrrrrr}
\toprule
Run & Conditions & Episodes & Budget & At budget & Max tokens & GPU time \\
\midrule
Qwen3-8B, main (chat) & 3 think + direct & 500 & 4{,}000 & 2.7--6.9\% & 13{,}571 & 5.4\,h \\
Qwen3.5-9B, main (chat) & 3 think + direct & 300 & 1{,}200 & 2.0--4.8\% & 7{,}752 & 0.8\,h \\
Qwen3-8B, plain text & 2 think + direct & 300 & 1{,}200 & 1.8\% / 6.1\% & 6{,}972 & 0.3\,h \\
Qwen3-8B-Base, plain text & 2 think + direct & 300 & 400 & $<$1\% & 1{,}408 & 0.1\,h \\
Qwen3-0.6B to 14B (chat) & think/body + direct & 300 & 4{,}000 & 0.5--6.9\% & 4{,}508 & 2.1\,h \\
Qwen3-8B, 16 turns (chat) & 3 think + direct & 300 & 4{,}000 & 1.5--5.4\% & 16{,}225 & 7.3\,h \\
Qwen3-8B, goal file (chat) & think/body & 300 & 4{,}000 & 12.4\% & 4{,}488 & 1.2\,h \\
\midrule
Qwen3-8B, note pool & think/full & 1{,}000 & 4{,}000 & 2.8\% & 16{,}083 & 4.5\,h \\
Qwen3.5-9B, note pool & think/full & 900 & 1{,}200 & 1.9\% & 8{,}112 & 0.8\,h \\
Qwen3.5-9B, pair pool & think/body & 300 & 1{,}200 & 2.8\% & 1{,}736 & 0.2\,h \\
\bottomrule
\end{tabular}
\end{table}

\begin{table}[!ht]
\centering\small
\caption{Experiments on saved histories (greedy decoding). Prompts $\times$ cells: saved prompts (bases for the update experiments) times the cells evaluated on each; for counterfactual pairs, the number of pairs (both versions of each toggled or reference pair are counted as prompts). Budget: maximum tokens of generated reasoning. Max tokens: the longest prompt plus generated reasoning. GPU time: wall time on one RTX A6000, Qwen3-8B / Qwen3.5-9B. Experiments without generated reasoning have no budget.}
\label{tab:configs_saved}
\footnotesize
\setlength{\tabcolsep}{3.5pt}
\begin{tabular}{llrrr}
\toprule
Experiment (appendix) & Prompts $\times$ cells & Budget & Max tokens & GPU time \\
\midrule
Fixed history (\ref{app:fixed}) & 400 $\times$ 5 & 3{,}000 & 3{,}467 / 1{,}564 & 4.4 / 1.3\,h \\
Own reasoning regenerated, Qwen3-8B (\ref{app:fixed}) & 400 $\times$ 3 & 6{,}000 & 6{,}469 & 0.7\,h \\
Retention with and without current reasoning (\ref{app:retention}) & 400 $\times$ 6 & 6{,}000 & 13{,}435 / 8{,}311 & 1.4 / 0.5\,h \\
Content of retained notes (\ref{app:content}) & 400 $\times$ 7 & --- & --- & 0.1 / 0.3\,h \\
Counterfactual pairs, swap and edit (\ref{app:cf}) & 160 / 150 pairs & 3{,}000 & 3{,}468 / 1{,}170 & 2.9 / 0.6\,h \\
Pairs with a toggled lock (\ref{app:cftoggle}) & 488 / 1{,}212 $\times$ 3 & 6{,}000 & 6{,}470 / 6{,}356 & 0.9 / 0.9\,h \\
Notes before a state change (\ref{app:fresh}) & 150 $\times$ 8 & 6{,}000 & 7{,}021 / 6{,}875 & 0.7 / 0.2\,h \\
Explicit event labels (\ref{app:events}) & 150 $\times$ 8 & 6{,}000 & 7{,}073 / 6{,}875 & 0.4 / 0.2\,h \\
Probes and TF-IDF on toggled pairs (\ref{app:stale}) & --- & --- & --- & 0.1 / $<$0.1\,h \\
Activation patching (\ref{app:patching}--\ref{app:patching_toggle}) & --- & --- & --- & 4.1 / 6.1\,h \\
Steering, Qwen3-8B (\ref{app:steer}) & --- & --- & --- & 2.4\,h \\
\bottomrule
\end{tabular}
\end{table}

\end{document}